%% file: acl_latex.tex
\pdfoutput=1

\documentclass[11pt]{article}

\usepackage[]{acl}

\usepackage{times}
\usepackage{booktabs}
\usepackage{latexsym}
\usepackage{amssymb}
\usepackage{amsmath}
\usepackage{listings}
\usepackage{pifont}
\usepackage{enumitem}
\usepackage[T1]{fontenc}

\usepackage[utf8]{inputenc}

\usepackage{microtype}

\usepackage{inconsolata}
\usepackage{hyperref}
\usepackage{url}
\usepackage{wrapfig}
\usepackage{graphicx}
\usepackage{lipsum}
\usepackage{booktabs} 
\usepackage{microtype}
\usepackage{inconsolata}
\usepackage{amssymb}
\usepackage{booktabs}
\usepackage{hyperref}
\usepackage{colortbl}
\usepackage{caption}
\usepackage[most]{tcolorbox}
\usepackage{amsmath}
\usepackage{multirow}
\usepackage{color}
\usepackage{pifont}       
\usepackage{bbding}
\usepackage{pifont}
\usepackage{algorithm}
\usepackage{algpseudocode}
\definecolor{deepred}{HTML}{FC8D59}
\definecolor{medred}{HTML}{FDC8B0}
\definecolor{lightred}{HTML}{F5E1D9} 
\definecolor{verylightred}{HTML}{F5E1D9}
\definecolor{extremelylightred}{HTML}{FCF4EF}
\definecolor{lightblue}{HTML}{E4F0F7}
\definecolor{medblue}{HTML}{CBE1EE}
\usepackage{mdframed}
\definecolor{green}{HTML}{65A542}
\usepackage{color}
\definecolor{shadecolor}{rgb}{0.92,0.92,0.92}
\newlength{\contentwidth} 

\usepackage{hyperref}
\hypersetup{hidelinks,
	colorlinks=true,
	allcolors=black,
	pdfstartview=Fit,
	breaklinks=true}

\usepackage{xcolor}

\title{LLMs-as-Instructors: Learning from Errors \\ Toward Automating Model Improvement}

\author{Jiahao Ying\textsuperscript{1, 2}\thanks{\ \ This work was performed when Jiahao Ying was an Intern at Skywork Al.}, Mingbao Lin\textsuperscript{2}, Yixin Cao\textsuperscript{3}\thanks{\ \ Corresponding author.}, Wei Tang\textsuperscript{4}, Bo Wang\textsuperscript{5},  \\
 \textbf{Qianru Sun}\textsuperscript{1}\textbf{,} \textbf{Xuanjing Huang}\textsuperscript{3}\textbf{,} \textbf{Shuicheng Yan}\textsuperscript{2}$^{\dag}$\\    
        \textsuperscript{1}Singapore Management University, Singapore \\
\textsuperscript{2}Skywork AI, Singapore   \\
\textsuperscript{3}Fudan University, China   \\
        \textsuperscript{4}
        University of Science and Technology of China, China 
  \\ \textsuperscript{5}Beijing Institute of Technology, China
        }

\begin{document}
\maketitle

\begin{abstract}
This paper introduces the innovative ``LLMs-as-Instructors'' framework, which leverages the advanced Large Language Models (LLMs) to autonomously enhance the training of smaller target models. Inspired by the theory of ``Learning from Errors'', this framework employs an instructor LLM to meticulously analyze the specific errors within a target model, facilitating targeted and efficient training cycles. Within this framework, we implement two strategies: ``Learning from Error,'' which focuses solely on incorrect responses to tailor training data, and ``Learning from Error by Contrast,'' which uses contrastive learning to analyze both correct and incorrect responses for a deeper understanding of errors. Our empirical studies, conducted with several open-source models, demonstrate significant improvements across multiple benchmarks, including mathematical reasoning, coding abilities, and factual knowledge. Notably, the refined Llama-3-8b-Instruction has outperformed ChatGPT, illustrating the effectiveness of our approach. By leveraging the strengths of both strategies, we have attained a more balanced performance improvement on both in-domain and out-of-domain benchmarks. Our code can be found at \url{https://yingjiahao14.github.io/LLMs-as-Instructors-pages/}.

\end{abstract}

\input{chapters/intro_1}

\input{chapters/dataset}

\input{chapters/exp}
\input{chapters/related_work}
\input{chapters/conclusion}

\input{chapters/limination}

\bibliography{anthology,custom}
\input{chapters/appendix}


\end{document}

%% file: chapters/intro_1.tex
\section{Introduction}
\begin{figure}[t]
\centering
\includegraphics[scale=0.36]{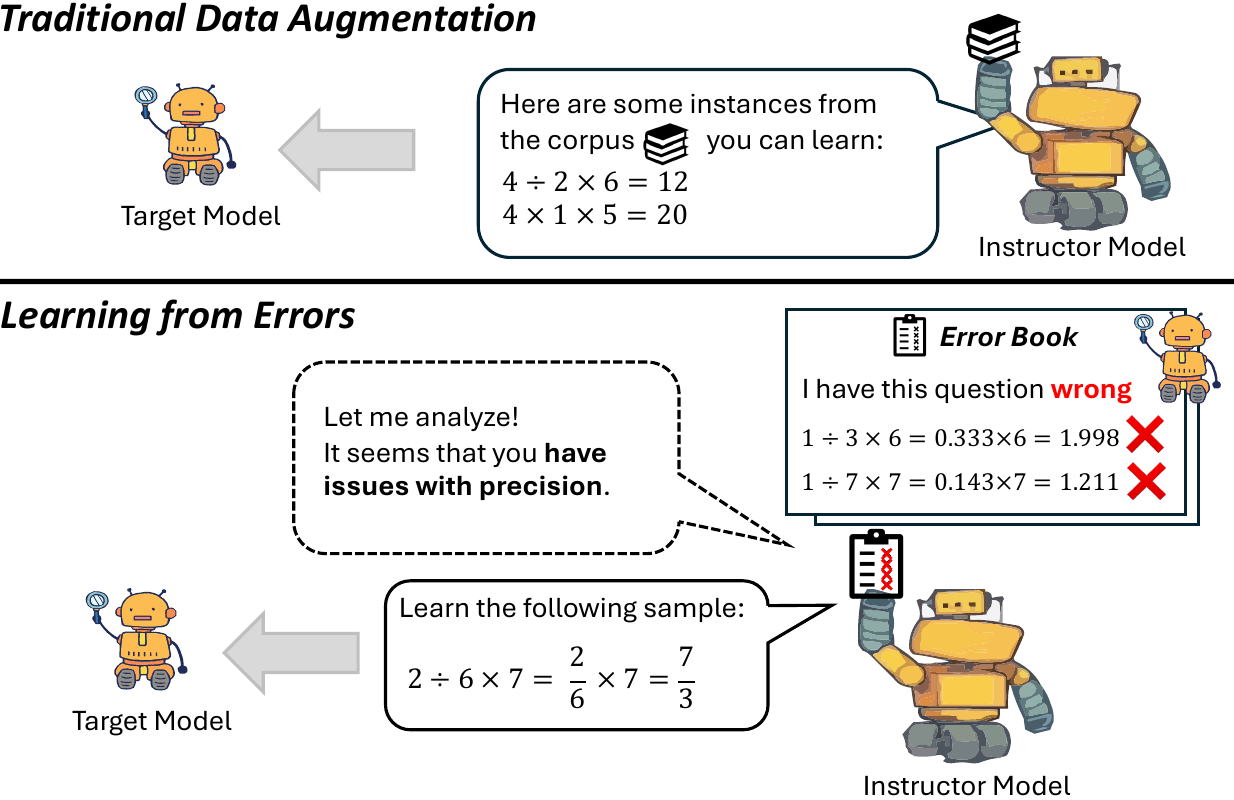}
    \caption{Compared with previous traditional data augmentation method~\cite{dai2023auggpt, xu2023wizardlm} --- used stronger models to generate training data without considering the target model's features, we propose an LLMs-as-Instructor framework based on the ``Learning from Errors'' theory~\cite{metcalfe2017learning}, where a stronger model analyzes and targets the specific errors of a smaller model to facilitate direct improvements.}
    \label{fig: intro}  
\vspace{-4mm}
\end{figure}

LLMs have achieved great success in various tasks, while it is still time-consuming and labor-intensive for further training, e.g., adapting to specific domains, following instructions, or aligning with human preference. To ensure continuous improvement during training, developers usually keep analyzing the model's responses and modifying/supplementing the corpus accordingly. Some advanced LLMs, such as GPT-4~\cite{GPT-4} and Claude-3~\cite{claude-3}, have been introduced to improve efficiency and reduce cost via data augmentation~\cite{dai2023auggpt} or labeling~\cite{rafailov2024direct}. As shown in Figure~\ref{fig: intro}, once LLMs are found poor performance in mathematical capability, more math data will be integrated deliberately into the next training cycle.

In this paper, we argue that conventional methods, which generate data indiscriminately to augment training corpus, are inefficient. They fail to fully leverage the capabilities of the advanced LLMs. Except for data generation, advanced LLMs actually also exhibit exceptional analytical and evaluative competencies~\cite{bai2023benchmarking,dai2023auggpt}. We are thus inspired by Bloom's theory~\cite{bloom19842}, which suggests that personalized tutoring is a highly effective educational approach for students,  to enhance LLMs' further training by mirroring the personalized and focused strategies in human education.

To do so,  we tailor advanced LLMs to serve as instructors to automate model improvement with minimal human intervention. The basic idea is that advanced LLMs first analyze those error-prone tasks and then augment training data to the point. That is, it would be more efficient by ``Learning from Errors" as discussed in~\cite{metcalfe2017learning}. This theory indicates that \textit{errorful learning followed by corrective feedback is beneficial to learning} for humans. As illustrated in Figure\,\ref{fig: intro}, different from conventional methods, we focus on those problematic responses. By analyzing specific errors, it is determined that the target model exhibited poor capacity in computational precision. Thus, numerous precision-focused questions are generated to enhance this capability in the next round of training.

In specific, our ``LLMs-as-Instructors'' has two strategies: \textbf{1) Learning from Error:} where the instructor model analyzes only the incorrect responses and corresponding question samples, facilitating identifying the underlying errors and generating tailored training data, \textbf{2) Learning from Error by Contrast:} building on the concept of contrastive learning, it involves both correct and incorrect samples. Correct samples closely related to each incorrect one will be paired for analysis. This strategy allows the Instructor to conduct a comprehensive analysis, enabling it to pinpoint the subtle distinctions between correct and incorrect responses.
%
%
 Our experiments, utilizing several open-source models serving as target models, have demonstrated that after improvements, our ``LLMs-as-Instructors'' framework significantly enhances the baseline performance of these target models across various benchmarks, which include testing mathematical reasoning with GSM8k~\cite{cobbe2021training}, coding abilities with HumanEval~\cite{chen2021evaluating}, and factual knowledge with MMLU~\cite{hendryckstest2021}.
The outcomes also reveal that each strategy offers unique benefits contigent upon the nature of the data. \textbf{1)} In contexts where the question samples are granular and there is an abundance of both correct and incorrect cases, the use of contrast samples substantially benefits the Instructor, \textbf{2)} On the other hand, for more general and less frequent questions, the ``Learning from error'' strategy is directly more efficacious in augmenting the model's capabilities. By combining the strengths of both strategies, we have realized a more balanced improvement in performance on the involved benchmarks.

Our main contributions can be summarized as:
\setlist[itemize]{left=0pt}
\begin{itemize}
    \item We have proposed the LLMs-as-Instructors framework, which leverages the capabilities of advanced LLMs to autonomously analyze the characteristics of the target model and enhance it through iterative training cycles.
    \item We have introduced and effectively implemented two distinct analysis strategies: ``Learning from Errors'' and ``Learning from Errors by Contrast,'' which are adaptable to various scenarios and facilitate learning from errors.
    \item  Our evaluation spans multiple benchmarks and diverse domains, including factual knowledge, mathematical reasoning, and coding, confirming that our framework significantly improves the performance of target models.
\end{itemize}

%% file: chapters/dataset.tex
\begin{figure*}[!t]
    \centering
    
    \includegraphics[scale=0.472]{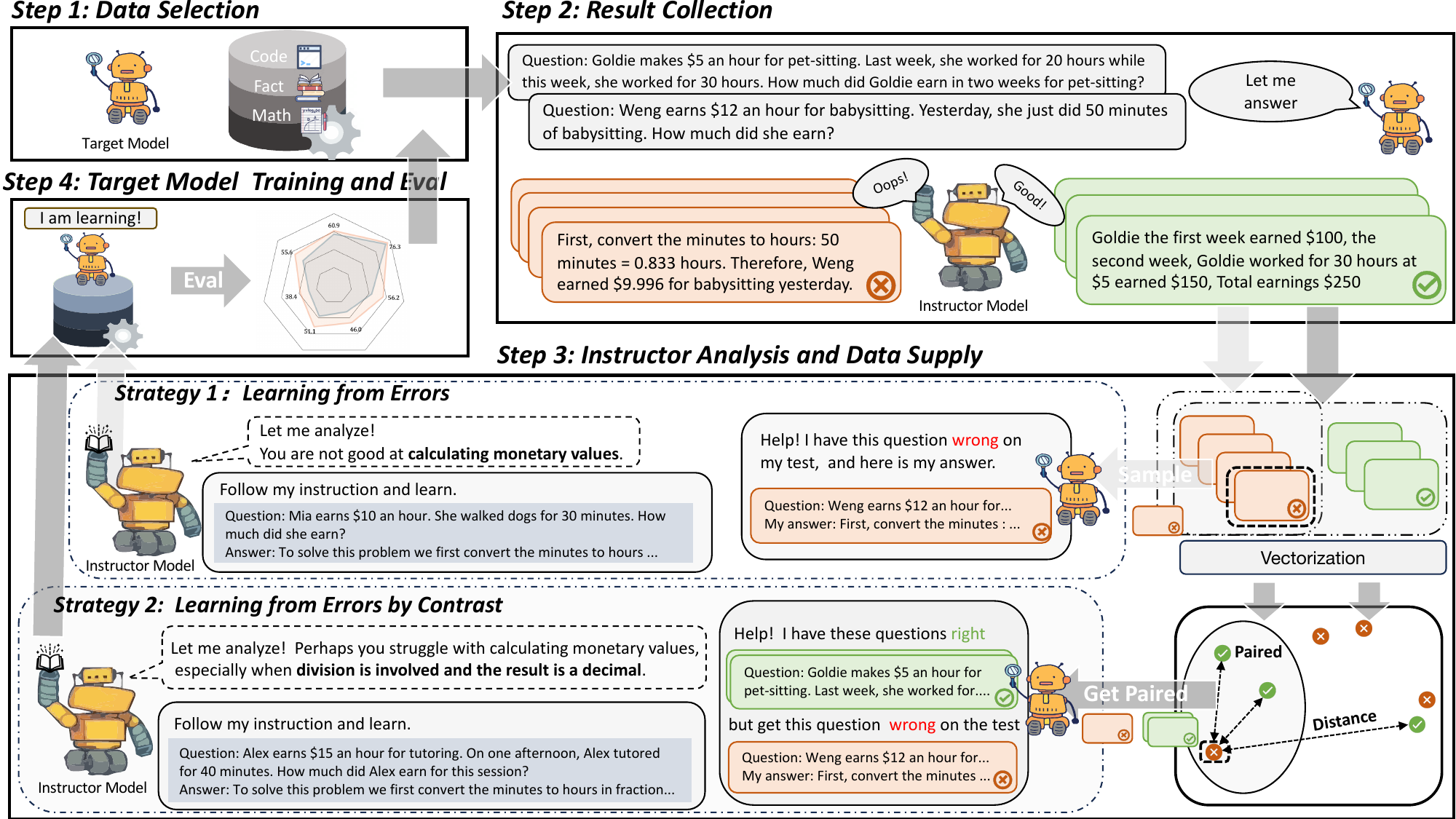}
   \caption{Our LLMs-as-Instructors framework consists of four steps in each iteration cycle to improve the target model: 1. Data Selection (Section~\ref{sec: evaluation data selection}), where target data samples are selected to challenge and assess the capabilities we intend to enhance. 2. Result Collection (Section~\ref{sec: evaluation data selection}), involving the evaluation of the target model on these samples and collection of responses for analysis. 3. Instructor Analysis and Data Supply (Section~\ref{sec: Instructor Analysis and Data Supply}), where the instructor conducts analysis and generates tailored training data. 4. Target Model Training and Evaluation (Section~\ref{Target Model Fine-Tuning and Evaluation}), having the target model learn from the errors and conducting the assessment of the improvements.}
    \label{fig:framwork}
\end{figure*}

\section{LLMs-as-Instructors Framework}

The framework depicted in Figure\,\ref{fig:framwork} illustrates our LLMs-as-Instructors approach, which is characterized by undergoing a series of $n$ iterations across four integral stages:
%
%
\textbf{1) Data Selection} (Section\,\ref{sec: evaluation data selection}) is the initial step where we meticulously select target data samples designed to evaluate the capabilities we intend to enhance in subsequent iterations.
%
%
\textbf{2) Result Collection} (Section\,\ref{sec: evaluation data selection}) follows, where we scrutinize the performance of the target model and meticulously gather its responses for in-depth subsequent analysis.
%
%
\textbf{3) Instructor Analysis and Data Supply}  (Section\,\ref{sec: Instructor Analysis and Data Supply}) is the phase where the instructor, employing a predefined analytical strategy, dissects the errors made by the target model. This analysis is pivotal in formulating strategies for improvement and crafting tailored training corpora to facilitate the target model's learning process.
%
%
\textbf{4) Target Model Training and Evaluation} (Section\,\ref{Target Model Fine-Tuning and Evaluation}) concludes the cycle, where the target model undergoes fine-tuning with the supplemented training corpus. It is then  evaluated to measure the extent of improvement using a dedicated evaluation dataset.

\begin{algorithm}[htb]
\small
\caption{Given a target model $\mathcal{M}^0_{\text{target}}$, target dataset including $\mathcal{D}_{\text{target}}$ and evaluation datasets $\mathcal{D}_{\text{eval}}$, we deploy the instructor model $\mathcal{M}_{\text{Instructor}}$. This instructor analyzes the responses from the target model $\mathcal{M}^i_{\text{target}}$ at each iteration $i$. Based on this analysis, it generates corresponding supplemental training corpora designed to iteratively enhance the target model’s performance.}
\label{algorithm: LLMs-as-Instructors}
\begin{algorithmic}[1]
\Procedure{LLMs-as-Instructors}{} 
\Statex{ \ \ \ \ \ \ \ \ \ \ \ \ ($\mathcal{M}^0_{\text{target}}, \mathcal{M}_{\text{Instructor}}$}{ $\mathcal{D}_{\text{target}}, \mathcal{D}_{\text{eval}}$)}
\State $i \gets 0$
\While{$i < n$}
    \State $\mathcal{D}_{\text{target}}^i\gets \text{DataSelect}(\mathcal{D}_{\text{target}}) $
    \State $R^i \gets \text{Evaluate}(\mathcal{M}^i_{\text{target}},\mathcal{D}_{\text{target}}^i, 
        \mathcal{M}_{\text{Instructor}})$
    \State $\mathcal{D}_{\text{train}}^i \gets \text{Generate}(\mathcal{M}_{\text{Instructor}},\mathcal{D}_{\text{target}}^i, R^i)$
    \State $\mathcal{M}^i_{\text{target}} \gets \text{Finetune}(\mathcal{M}^i_{\text{target}}, \mathcal{D}_{\text{train}}^i)$
    \State \text{Evaluate}( $\mathcal{M}_{\text{\text{target}}}^i, \mathcal{D}_{\text{eval}}, \mathcal{M}_{\text{Instructor}})$
    \State $i \gets i + 1$
\EndWhile
\EndProcedure
\end{algorithmic}
\end{algorithm}
Upon receiving a target LLM $\mathcal{M}^0_{\text{target}}$, we engage a stronger model as the ``Instructor'' ($\mathcal{M}_{\text{Instructor}}$), utilizing responses from the target model to analyze the errors, and then generating customized training materials to guide the model to learn from the errors for performance increase. The process of this method is outlined in Algorithm~\ref{algorithm: LLMs-as-Instructors}.

\subsection{Preliminary}

To simplify the subsequent discussion, we establish a set of definitions: The dataset $\mathcal{D}$ is defined as follows: $\mathcal{D} = \{d \mid d = (q, a_{\text{ref}})\}$, where each sample $d$ includes a question $q$ and a reference answer $a_{\text{ref}}$.
We designate an erroneous case as $(d^-, r^- )$ when $d^{-} = (q, a_\text{ref})$ and the target model's response \( r^- \) to question $q$ does not match \( a_\text{ref} \). Conversely, a correct case is denoted as $(d^+, r^+ )$.

\subsection{Data Selection and Result Collection}
\label{sec: evaluation data selection}
\textit{``If you can't measure it, you can't improve it.''} 
%
%
%
To initiate the model enhancement process, we first engage in the critical task of measuring and identifying the errors of target model. As outlined in line 4 of Algorithm\,\ref{algorithm: LLMs-as-Instructors}, this is achieved by carefully selecting a subset of the target dataset, denoted as $\mathcal{D}_{\text{target}}^i$, from the base $\mathcal{D}_{\text{target}}$.
Utilizing the chosen samples from $\mathcal{D}_{\text{target}}^i$, we proceed to evaluate the performance of the current iteration of the target model, $\mathcal{M}^i_{\text{target}}$, with the potential support the instructor model $\mathcal{M}_{\text{Instructor}}$, if required. This evaluation results in a collection of responses, denoted as $R^i$, as outlined in line 5 of Algorithm\,\ref{algorithm: LLMs-as-Instructors}.

Following this initial assessment, the instructor model delves deeper into the analysis of these responses, aiming to identify areas for improvement and to provide valuable insights for the subsequent enhancement of the target model in Section\,\ref{sec: Instructor Analysis and Data Supply}. 
\subsection{Instructor Analysis and Data Supply}
\label{sec: Instructor Analysis and Data Supply}

Subsequently, the instructor $\mathcal{M}_\text{Instructor}$ further leverages the collected responses \( R^i \) and applies the following two strategic analyses to aid the target model $\mathcal{M}^i_\text{target}$ in learning from its errors.

\setlist[itemize]{left=0pt}
\begin{itemize}
   \item \textbf{Learning from Errors (LE):}
   The instructor $\mathcal{M}_{\text{Instructor}}$ focuses the analysis solely on the incorrect responses $r^{-}$ from $R^i$, along with the corresponding target sample 
   $d^{-}$ from $\mathcal{D}_{\text{target}}^i$. The training dataset  $\mathcal{D}_{\text{train}}^i$ is then generated to help the target model learn from its errors. In this context, Algorithm\,\ref{algorithm: LLMs-as-Instructors}, line 6, is adapted to:
\begin{equation}
   \begin{aligned}
       &\mathcal{D}_{\text{train}}^i \gets \text{Generate}(\mathcal{M}_{\text{Instructor}},\mathcal{D}_{-}^i),
   \end{aligned}
\end{equation}
where \scalebox{0.8}{$\mathcal{D}_{-}^i = \{(d^-, r^-) \mid d^- \in \mathcal{D}_{\text{target}}^i, r^- \in R^i\}$}. 
Following a set of instructions outlined in Appendix\,\ref{prompt llms-as-instructors}, the instructor model $\mathcal{M}_{\text{Instructor}}$ scrutinizes the current state of the target model $\mathcal{M}_{\text{target}}$. It then crafts tailored training materials $\mathcal{D}_{\text{train}}^i$ aimed at enhancing the target model's performance. This process is primarily driven by the identification and rectification of errors.

\item \textbf{Learning from Errors by Contrast (LEC):}
Beyond the erroneous cases \( (d^{-}, r^{-}) \), inspired by ``Contrastive Learning''~\cite{hadsell2006dimensionality, chen2020simple}, which highlights learning by comparing negative and positive samples, we incorporate correct cases \( (d^{+}, r^{+}) \) for contrast to enhance learning from errors. Specifically, for each erroneous case \( (d^{-}, r^{-}) \), the vectorized features \( \mathbf{v}(d^{-}, r^{-}) \) are used to calculate and retrieve the \( k \) most similar correct cases:
\begin{equation}
\resizebox{0.85\hsize}{!}{$
\begin{aligned}
   &\mathcal{D}_{\text{paired}}^{(d^-, r^-)} =\left\{  \underset{(d^{+}, r^{+})\in \mathcal{D}_{+}^{i}, k}{\operatorname{argmin}} \|\mathbf{v}(d^{-}, r^{-}) - \mathbf{v}(d^{+}, r^{+}) \|_2\right\},
\end{aligned} $}
\label{get sim}
\end{equation}
where \scalebox{0.8}{\( \mathcal{D}_{+}^i = \{(d^+, r^+) \mid d^+ \in \mathcal{D}_{\text{target}}^i, r^+ \in R^i\} \)}. These \( k \) retrieved paired cases, along with the incorrect case, form the contrast set. This set is provided to the instructor for detailed comparative analysis and to generate a training dataset that specifically targets the errors. Consequently, the procedure outlined in line 6 of Algorithm\,\ref{algorithm: LLMs-as-Instructors} is modified as:
\begin{equation}
\begin{aligned}
   \mathcal{D}_{\text{train}}^i  \gets \text{Generate}(\mathcal{M}_{\text{Instructor}},\mathcal{D}_{\text{paired}}^i), 
\end{aligned} 
\end{equation}
in which \resizebox{.8\hsize}{!}{$D_{\text{paried}}^i  = \left\{({d}^{-}, {r}^{-}, p )\mid (d^{-}, r^{-}) \in  \mathcal{D}_{-}^{i}, p \in \mathcal{D}_{\text{paried}}^{({d}^-, {r}^-)}   \right\}$}.
  
\end{itemize}

Note during the training process, the selection of strategies is not static; it is amenable to modification in accordance with the evolving state of the target model. We will delve deeper into this aspect of adaptability in our experimental Section\,\ref{Differential Impact of Analysis Strategies on Iterative Improvement}.

\subsection{Target Model Training and Evaluation}
\label{Target Model Fine-Tuning and Evaluation}
Using the supplemental training data $\mathcal{D}_{\text{train}}^i$ from the instructor model, the target model is fine-tuned to learn from its errors (line 7 of Algorithm\,\ref{algorithm: LLMs-as-Instructors}). To fairly evaluate the improvements after training, we utilize additional evaluation datasets, $\mathcal{D}_{\text{eval}}$, considering that using the same datasets for both training and evaluation can lead to an overestimation of the model’s performance~\cite{ying2024automating, zhou2023don} (lines 8 of Algorithm~\ref{algorithm: LLMs-as-Instructors}).  After evaluating the enhancements, we can proceed with further iterations of learning. This iterated learning process allows for continual refinement.

%% file: chapters/exp.tex
\section{Experimentation}\label{experiment}

\subsection{Experimental Datasets}\label{dataset}
Following~\cite{wang2023far}, we focus on three practical abilities: factual knowledge, mathematical reasoning, and coding.
We select a relevant public dataset for each, amalgamating their training data into the target dataset $\mathcal{D}_\text{target}$, and their test data into the evaluation dataset $\mathcal{D}_\text{eval}$. Given the similarity in the distributions of in-domain training and testing sets, learning from errors potentially inflates performance. 
Thus, for each ability, we also incorporate an ``Out-of-Domain'' (OOD) dataset in $\mathcal{D}_\text{eval}$.
This setup allows us to assess the model’s ability to generalize and enhance its performance beyond its initial In-Domain (ID) training context.
\setlist[itemize]{left=0pt}
\begin{itemize}
\item \textbf{For Factual Knowledge}, we select Massive Multitask Language Understanding dataset (MMLU)~\cite{hendryckstest2021}, with questions across 57 subjects, as the ID dataset. 
For OOD evaluation, we use the ARC-Challenge dataset~\cite{allenai:arc}, which comprises science questions that also test factual knowledge.
\item \textbf{For Mathematical Reasoning}, we select GSM8k~\cite{cobbe2021training}, a Grade School Math dataset, as our ID dataset. For OOD evaluation, we use GSM8k-PLUS~\cite{li2024gsm}, an extension of GSM8k augmented with various mathematical perturbations.
\item \textbf{For Coding}, we use the MBPP~\cite{austin2021program} as the ID dataset. For OOD evaluation, we select HumanEval~\cite{chen2021evaluating}. 
\end{itemize}

\begin{table*}[!t]
\centering
\resizebox{0.9\textwidth}{!}{%
\begin{tabular}{lcc|cc|cc|cc}
\toprule
& \textcolor{green}{\textbf{MMLU}} & \textbf{ARC$_{c}$} & \textcolor{green}{\textbf{GSM8k}} & \textbf{GSM-PLUS} & \textcolor{green}{\textbf{MBPP}} & \textbf{HumanEval} & \textbf{BBH} &  \textbf{\textbf{Average}}\\
& \textbf{(factuality}) & \textbf{(factuality)} & \textbf{(math)} & \textbf{(math)} & \textbf{(coding)} & \textbf{(coding)} & \textbf{(holistic)} \\
\cmidrule{2-9}
& EM & EM & EM & EM & P@1 &P@1 & EM \\
& (0-shot) & (0-shot) & (0-shot) & (0-shot) & (0-shot) & (0-shot) & (3-shot) & \\
\midrule
\textit{Closed-source model and Open-source models} \\

GPT-3.5-turbo &  65.3 & \textbf{\underline{84.4}} & \textbf{71.1} & \textbf{61.5}& \textbf{\underline{68.5}} &\textbf{\underline{61.6}} & \textbf{61.7} & \textbf{67.7}\\
Gemma-7B-Instruct & 50.4 &72.8 & 35.7 & 27.0 &34.3 & 25.0& 44.8 &41.4 \\
Llama-2-70b-Chat &  60.0 & 77.2 & 51.0 & 42.0& 41.1 & 36.0 & 49.6 & 51.0\\
Mixtral-8$\times$7B-Instruct-v0.1  &\textbf{\underline{67.2}}&81.5 &50.4 & 41.4 & 27.9& 36.0 & 44.9 & 49.9\\
\midrule
\textit{Mistral-7b-Instruct-v0.2 based} \\
Vanilla & 57.3 & 73.9 & 43.2 & 32.5 & 38.0 & 28.6 & 44.3 & 45.4\\
+ Fine-tuning & 59.8 & 74.5 &\textbf{58.0} &43.6  & 31.5&22.6&50.4 & 48.6\\
+ AugGPT & 58.9 & 76.0 & 54.4 & 44.0 & 44.1 & 37.8 & 53.4 & 52.6\\

+ LLMs-as-Instructors (LE) & 60.9 & 76.3 & 56.2 & \textbf{46.0} & \textbf{51.5} & 38.4 & 55.9 & 55.0\\
+ LLMs-as-Instructors (LEC) & \textbf{61.5} & \textbf{77.2} & 54.0 & \textbf{46.0} & 47.0 & \textbf{43.3} & \textbf{56.8} & \textbf{55.1}\\
\midrule
\textit{Llama-3-8b-Instruction based} \\

Vanilla &  66.3 & 82.2 & 79.2 & 64.6 & 56.1 & 59.7 & 65.5 & 67.7\\
+ Fine-tuning & 66.3 & 82.3 & 79.5 & 64.6 & 55.1& 59.8 & 65.5 & 67.6\\
+ AugGPT & \textbf{66.5} & \textbf{83.0} & 77.8 & 64.1 & 56.1& 56.7 & 65.3& 67.1\\
+ LLMs-as-Instructors (LE) & 66.1& 82.5 & \textbf{\underline{81.2}} & 65.9 & \textbf{56.5}& 60.4 & 65.8 & 68.3\\
+ LLMs-as-Instructors (LEC) & 66.2& 82.8 & 80.2 &\textbf{\underline{66.0}} & \textbf{56.5}& \textbf{\underline{61.6}} & \textbf{\underline{66.1}} & \textbf{\underline{68.5}}\\
\bottomrule
\end{tabular}}
\caption{Performance (\%) of Mistral-7b-Instruct after three iterations of improvement under the LLMs-as-Instructors framework and Llama-3-8b-Instruction after one iteration, across seven selected benchmarks. The benchmarks highlighted in \textcolor{green}{green} denote the ID datasets. \textit{LE} denotes the exclusive use of analysis strategy: Learning from Errors, in all three iterations. \textit{LEC} denotes the use of analysis strategy: Learning from Errors by Contrast. \textit{AugGPT} denotes following work~\cite{dai2023auggpt, li2024common} to generate augmented samples (a total of 27,000) for training detailed in Section\,\ref{experiment setup}. \textbf{Bold} indicates the best in each setting, \textbf{\underline{bold underline}} indicates the best in the table.}
\label{tab: main result}
\end{table*}

We also utilize BIG-bench Hard (BBH)~\cite{suzgun2023challenging}, a holistic benchmark containing 23 challenging tasks from Big-Bench~\cite{srivastava2023beyond}, to evaluate the models' overall general reasoning capabilities. For those ID datasets that lack an official training set, we use the validation datasets instead. For time considerations, we have limited the size of GSM8k-PLUS to match that of GSM8k.  Statistical result for the selected benchmarks is shown in Table\,\ref{table: statistic detail} and detailed task descriptions are shown in Appendix\,\ref{benchmark details}.

\subsection{Experimental Setups}\label{experiment setup}
We use Mistral-7b-Instruct~\cite{mistral-7b} and Llama-3-8b-Instruction~\cite{Llama3} as the target model \scalebox{0.8}{$\mathcal{M}^0_\text{target}$}, with GPT-4-preview~\cite{GPT-4} as the instructor model \scalebox{0.8}{$\mathcal{M}_\text{Instructor}$}:

For \textit{Data Selection}, we pick training data from the three ID datasets for each round; for \textit{Result Collection}, except for BBH that uses 3-shot examples from the original benchmark, all other responses are generated in a zero-shot manner, details of which are provided in Appendix\,\ref{model setting}. For these benchmark-based assessments, we follow their standard metrics elaborated in Appendix\,\ref{benchmark details}.

For \textit{Instructor Analysis and Data Supply}, we employ BERT embeddings~\cite{devlin-etal-2019-bert} to transform the erroneous cases into vectors and apply $\ell_2$ distance to get paired cases in Equation\,\ref{get sim}. Recognizing that the examination of question features is more prevalent in human analysis, we utilize the feature vector of the question $q$ from ${d}^{-}$, denoted as $\mathbf{v}({q})$, in lieu of $\mathbf{v}({d}^{-}, r^{-})$. To ensure the quality of the pairings, we stipulate $d^-$ and $d^+$ must hail from the same source dataset and set the parameter $k =3$ within Equation\,\ref{get sim}. We regulate the volume of the generated training dataset $D^i_{train}$ to a cap of 9,000, thereby allocating 3,000 samples to each of the three target datasets per iteration.

As for the phase of \textit{Target Model Training and Evaluation}, we conduct experiments on 16 NVIDIA A800 GPUs in a full parameter fine-tuning settings. Details of training configurations are provided in Appendix\,\ref{model setting}.

\subsection{Experimental Baselines}
In addition to closed-source ChatGPT~\cite{chatgpt} (GPT-3.5-turbo), we also incorporate a range of open-source models including Gemma-7b-Instruct~\cite{Gemma}, Llama-2-70b-Chat~\cite{llama2}, and Mixtral-8$\times$7B-Instruct-v0.1~\cite{mixtral-moe}. Furthermore, we have integrated the data augmentation method, which employs more advanced models to aid in training, without considering the target model's features. Specifically, we have 
1. \textit{Fine-tuning}, where we use the training part of the target (Section\,\ref{dataset}) to train the target model. 
2. \textit{AugGPT}, here we use sample $D_{target}^i$ as a seed to generate training data through the instructor, following previous work~\cite{dai2023auggpt, li2024common}, but without any filtering (prompt is in Appendix\,\ref{prompt auggpt}).


\subsection{How effective is the LLMs-as-Instructors framework for LLMs improvement?}
Utilizing GPT-4-preview as the instructor, we have successfully enhanced both the Mistral-7b-Instruct and Llama-3-8b-Instruction models, employing two distinct analysis strategies. This enhancement is achieved through three iterative training phases for Mistral-7b-Instruct and one iterative phase for Llama-3-8b-Instruction, due to time and cost considerations. The final model's accuracy on seven selected benchmarks is shown in Table\,\ref{tab: main result}. The outcomes underscore several key points: 

\textbf{1)} The two models realize significant enhancements on both ID and OOD benchmarks, irrespective of the strategy employed. Specifically, the Mistral-7b-Instruct achieves average improvements of 9.4\% and 9.5\% for the two strategies, and Llama-3-8b-Instruction shows gains of 0.6\% and 0.8\%. These results substantiate the efficacy of our approach.
\textbf{2)} When juxtaposed with the two data augmentation methods, our LLMs-as-instructors strategy exhibits a pronounced advantage, outperforming all seven benchmarks with an average lead of 2.5\% over AugGPT augmentation and 6.4\% over fine-tuning for Mistral. Similarly, for Llama3, the improvements are 1.3\% and 0.7\%, respectively. This aligns with our expectations, as our customized approach, which is sensitive to the model's unique characteristics, emerges as a more potent learning mechanism.
\textbf{3)} In comparison with open-source models of similar or even larger scale, such as Gemma-7B-Instruct, Llama-2-70b-Chat, and Mixtral-8$\times$7B-Instruct, the refined Mistral-7b-Instruct demonstrates a clear superiority, particularly in coding, mathematical reasoning, and overall general reasoning. Both strategies surpass these models by an average of 14\%. Moreover, the refined Llama-3-8b-Instruction surpasses the performance of ChatGPT by 0.7\% averagely, achieving state-of-the-art results as shown in the table. Considering that the performance of Llama-3-8b-Instruction is already high, the fact that our framework could further enhance it underscores the effectiveness of our approach.

\section{Discussion}
Given the proven effectiveness of our LLMs-as-Instructors method, we further delve deeper to ascertain the various contributing factors to its success. Specifically, we have examined the impact of the training set size (Section\,\ref{Impact of Training Set Size on Improvement}), and the number of training iterations (Section\,\ref{Impact of Iterations on Improvement}), the distinct impacts of the two analysis strategies on iterative improvement in Section\,\ref{Differential Impact of Analysis Strategies on Iterative Improvement}. Furthermore, we have conducted preliminary experiments aimed at amalgamating the two strategies to achieve a balanced improvement (Section\,\ref{Differential Impact of Analysis Strategies on Iterative Improvement}).
Due to contextual and time constraints, this discussion focuses only on the impact of these variables on the Mistral-7b-Instruct. The results concerning the Llama-3-8b-Instruction model are detailed in Appendix~\ref{more experiment results}.

\subsection{Impact of Training Set Size}\label{Impact of Training Set Size on Improvement}

\begin{figure}[!t]
    \centering
    \includegraphics[scale=0.46]{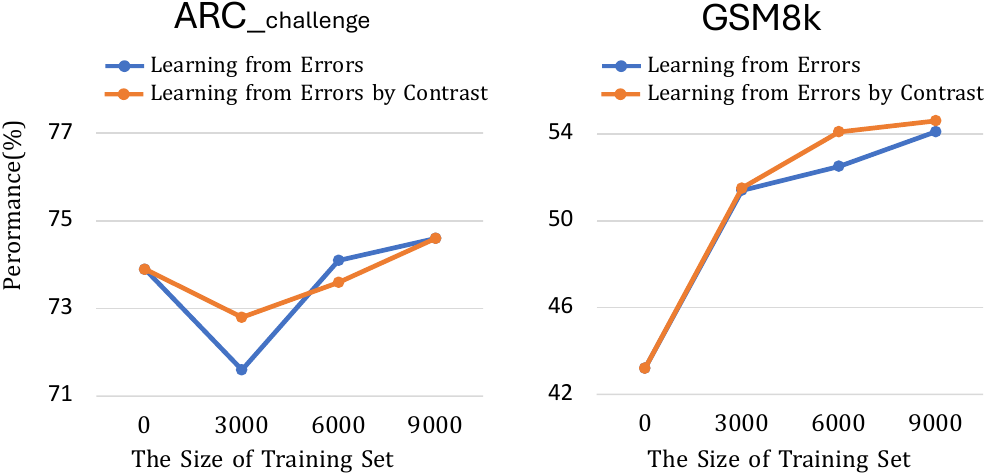}
    \caption{Performance (\%) of Mistral-7b-Instruct after the first iteration of improvement under the LLMs-as-Instructors framework, varied by the size of the training set. We compare three different training set sizes: 3,000, 6,000, and 9,000 total samples.}
    \label{Impact of Training Set Size}
   \vspace{-2mm}
\end{figure}

In this experiment, we elucidate the rationale behind selecting a trainign set size of 9,000 for our experimental configuration. Employing consistent parameters, we train the target model, Mistral-7b-Instruct, across a range of training set sizes, as depicted in Figure\,\ref{Impact of Training Set Size} (with more results presented in Figure\,\ref{Impact of Training Set Size all} and Figure\,\ref{Impact of Training Set Size llama3}). In the case of using the Learning from Error by Contrast strategy, preliminary experiments reveal that $k=3$ provided the best performance on the in-domain datasets (detail is shown in Figure~\ref{k number}). Therefore, we set 
$k$ to 3 for our quantitative experiments. Our observations indicate that the model's performance on ARC$_{c}$ only commence to escalate once the aggregate number of training samples surpasses the threshold of 6,000.
Moreover, as the sample size expands from 6,000 to 9,000, the enhancements in performance on the GSM8k benchmark, particularly with strategy LEC, begin to level off. Thus, we opt for a training set size of 9,000 samples to conduct improvement iteratively.

\subsection{Impact of Iterations on Improvement}\label{Impact of Iterations on Improvement}

\begin{figure*}[!t]
    \centering   
    \includegraphics[scale=0.45]{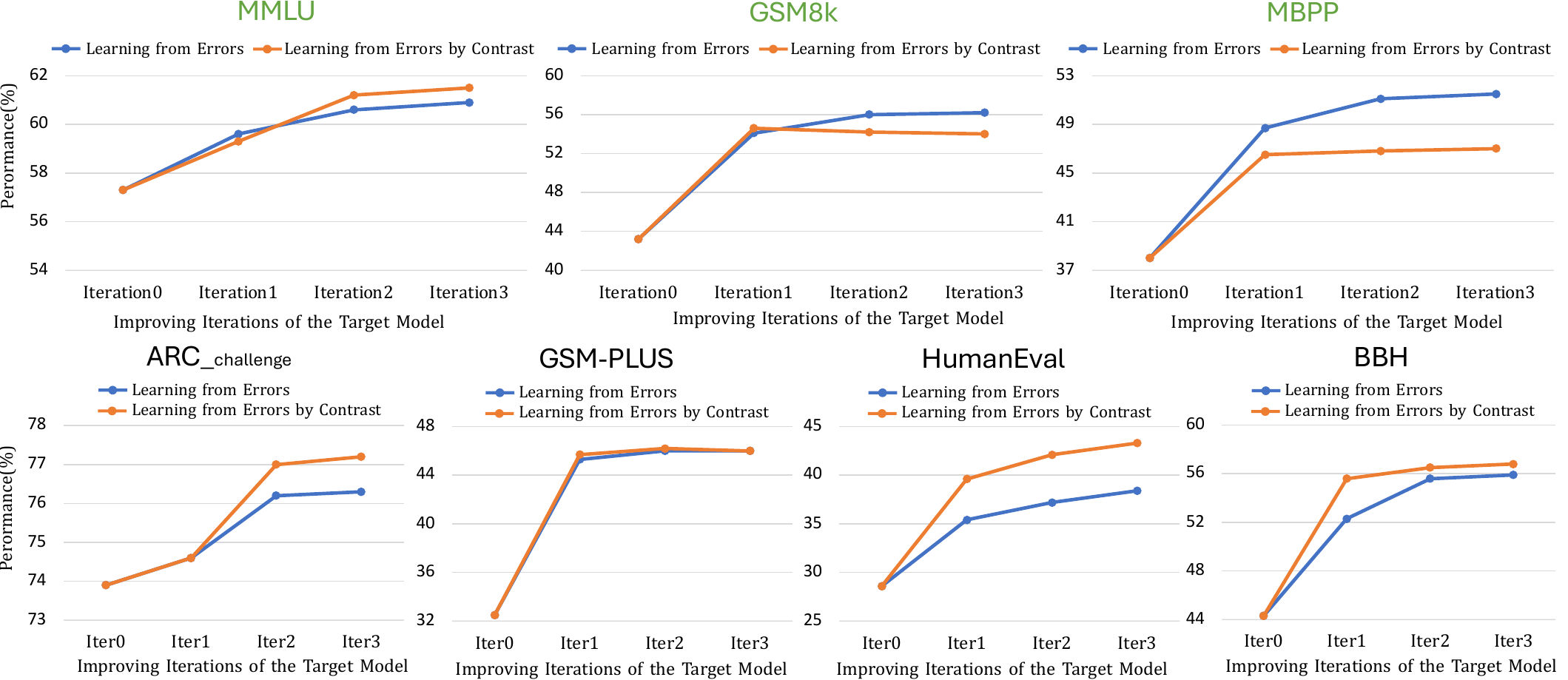}
    \caption{Performance (\%) of  Mistral-7b-Instruct under the LLMs-as-Instructors framework across different numbers of iterations. The benchmarks highlighted in \textcolor{green}{green} denote the ID datasets.}
    \label{Impact of Iterations on Improvement Fig}
\end{figure*}

\begin{figure}[!t]
    \centering
    \includegraphics[scale=0.65]{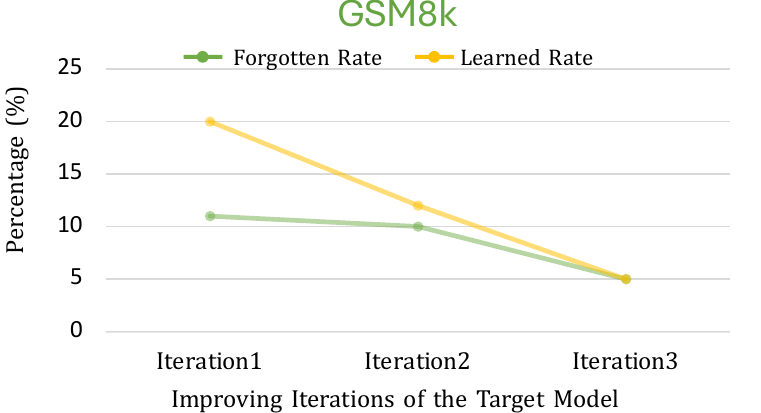}
    \caption{The target model's learning and forgetting behaviors evolving with successive iterations using strategy 1: Learning from Errors.}
    \label{learning and forgetting behaviors}
    \label{T}
\vspace{-2mm}
\end{figure}

We continue to investigate how the model iteratively improves. Through three successive training iterations of Mistral-7b-Instruct, we present the outcomes for each iteration in Figure\,\ref{Impact of Iterations on Improvement Fig}. Our findings reveal that under both strategies,  the model consistently achieves either progressive improvement or sustains a performance level on par with the preceding iteration in each round. However, as the iterations accumulate, the rate of enhancement tends to diminish. To delve deeper into the underlying causes, we examine the evolution of the model's learning and forgetting dynamics across consecutive iterations. Utilizing the GSM8k dataset, we calculate the Forgotten Rate and Learned Rate for iteration $i$ where $i >1$ as follow:
\begin{equation}
\resizebox{0.8\hsize}{!}{$
\begin{aligned}
  &Forgotten \ Rate ^{i} = \frac{|\mathcal{D}^{i-1}_{eval+} -\mathcal{D}^{i}_{eval+} |}{|\mathcal{D}_{eval}|},  \\
   &Learned \ Rate ^{i} = \frac{|\mathcal{D}^{i}_{eval+} -\mathcal{D}^{i-1}_{eval+} |}{|\mathcal{D}_{eval}|},  \\
\end{aligned} $}
\end{equation}
where \scalebox{0.8}{$\mathcal{D}^{i}_{eval+} $} denotes samples that the target model \scalebox{0.8}{$\mathcal{M}^{i}_{target}$} correctly addresses. The results, as illustrated in Figure\,\ref{learning and forgetting behaviors}, demonstrate that with an increasing number of iterations, both the learned rate and the forgotten rate of the model exhibit a decline. Notably, the decrease in the learned rate is more pronounced than that in the forgotten rate, leading to a nearly balanced state between learning and forgetting in the third iteration. This learning pattern contributes to the diminishing marginal gains in model performance as the iterations continue. This also suggests that increasing the initial learning rate or mitigating the forgotten issues during the learning process might further improve performance.

\subsection{Impact of Analysis Strategies on Iterative Improvement}\label{Differential Impact of Analysis Strategies on Iterative Improvement}
The analysis reveals that the two strategies exhibit different patterns of improvement. As depicted in Figure\,\ref{Impact of Iterations on Improvement Fig}, focusing solely on error cases for learning from mistakes yields better results in in-domain math and coding skills, with increases of 1.8\% and 4.5\% respectively, compared to using contrast sets. However, for the MMLU benchmark, contrastive pairs perform better. This discrepancy may stem from the contrast set quality, \scalebox{0.8}{$\mathcal{D}_{\text{paired}}^{(d^-, r^-)}$}. The MMLU's fine-grained categorization of subjects into various disciplines means that while similar correct and incorrect pairs are formed, they may not cover the same knowledge domains, thus not enhancing the comprehensiveness of the analysis.
This indicates the varied applicability of the strategies --- Learning from Error by Contrast is advantageous in scenarios with a rich set of question samples, while Learning from Errors is more effective for general questions.
Also, OOD evaluations show an average 1.7\% performance boost with contrastive pairs, indicating that it produces more generalizable data.

Leveraging the unique strengths of both strategies, we undertake preliminary experiments to enhance the model's performance more evenly:

1. \textit{Sequential Application of Strategies}. Acknowledging strategy LE's advantage in mathematical and coding tasks within in-domain contexts, we apply it to the target model \( M^2_{\text{target}} \), which had already undergone two rounds of improvement with strategy LEC. In the subsequent training round, we observe further progress, as in Table\,\ref{preliminary experiment for combine two stratigies}. Notably, the model sustains its OOD capabilities while achieving a 1.9\% increase in ID coding proficiency.

2. \textit{Capitalizing on Strategy LEC's Generalizability}. Benefiting from the enhanced generalizability of data generated by strategy LEC, we implement an extra round of improvement on the target model \( M^2_{\text{target}} \), which has been previously augmented by Strategy LE. The results in Table\,\ref{preliminary experiment for combine two stratigies} reveal that this refinement maintains the model's ID performance while also securing a 0.6\% advancement in OOD coding. More results are provided in Table\,\ref{preliminary experiment for combine two strategies all}.

\begin{table}[htb]
\centering
\resizebox{0.46\textwidth}{!}{%
\begin{tabular}{lcc|cc}
\toprule
&  \textcolor{green}{\textbf{GSM8k}} & \textbf{GSM-PLUS} & \textcolor{green}{\textbf{MBPP}} & \textbf{HumanEval} \\
& \textbf{(math)} & \textbf{(math)} & \textbf{(coding)} & \textbf{(coding)}\\
\cmidrule{2-5}
 & EM & EM & P@1 &P@1  \\
\textit{Mistral-7b}  & (0-shot) & (0-shot) & (0-shot) & (0-shot) \\
\midrule
+ LaI ({LEC})  & 54.0 & 46.0 & 47.0 & 43.3  \\
+ LaI ({LEC + LE}) & 53.9 & 45.7 &\cellcolor{verylightred} 48.9 & \cellcolor{lightblue}42.7 \\
\midrule
+ LaI ({LE})  & 56.2 & 46.0 & 51.5 & 38.4 \\
+ LaI ({LE + LEC}) &\cellcolor{lightblue}{55.9} & \cellcolor{verylightred}46.1 & 51.5 & \cellcolor{verylightred}39.0 \\
\bottomrule
\end{tabular}}
\caption{\textit{LaI ({LEC + LE})} denotes using strategy LEC for the first two iterations and strategy LE for the third iteration using LLMs-as-Instructors. \textit{LaI ({LE + LEC})} denotes the application of strategy LE for the first two iterations and strategy LEC for the third iteration. \textit{LaI (LEC)}, and \textit{LaI (LE)} donate using strategy LEC and strategy LE for three iterations respectively. Blue cells indicate reduced performance, Red cells indicate improved performance.}
\label{preliminary experiment for combine two stratigies}
\vspace{-4mm}
\end{table}

\section{Generated Training Sample Analysis}
Results indicate that the model achieves substantial improvements on both in-domain and out-of-domain datasets, thereby affirming the generalizability of our method. In this section, we conduct a quantitative analysis of the characteristics of the training data generated to further demonstrate that these performance enhancements are not merely due to similarities between the training and evaluation data. For each generated training sample, we calculate the closest match within the evaluation dataset $D_{eval}$ using ROUGE-L~\cite{lin-2004-rouge} score. The results, presented in Table~\ref{Similarity score}, indicate that the characteristics of the generated training data are not closely aligned with those of the test data, with an average ROUGE-L score of 0.30.

%% file: chapters/related_work.tex
\section{Related Work}

\textbf{Instructor Model Guided Model Improvement:} 
Enhancing a target model is a widely adopted methodology that leverages the capabilities of LLMs to create innovative training datasets~\cite{dai2023auggpt, xu2023wizardlm, mitra2024orca, wang2024far, ivison2023camels, chen2023alpagasus, mitra2023orca, fu2023specializing, kumar2020data, li2024common, li2023tuna}. 
Concurrent with this, another study~\cite{lee2024llm2llm} employs an ``Instructor'' model to facilitate learning from the target model's errors. However, this method trains individual models for each dataset, thereby overlooking the potential advantages of integrating correct samples into the training.

\textbf{Self-Improving LLMs:} 
The current landscape of research sees several endeavors exploring the concept of self-improvement within models~\cite{tong2024can, tang2024,wu2023self, madaan2024self, wang2023self, yuan2024self, sun2024principle, burns2023weak, li2023self}. While these studies predominantly concentrate on autonomous iterations without external influences, our approach diverges by incorporating feedback mechanisms. These mechanisms not only capitalize on the target models' errors but also introduce direct intervention through strategic analysis. Such an integration mainly aims to refine the model's learning process, thereby fostering a more sophisticated and dynamic model evolution.

%% file: chapters/conclusion.tex
\section{Conclusion}

In this work, we introduced the innovative ``LLMs-as-Instructors'' framework, leveraging advanced Large Language Models (LLMs) to guide smaller target models in learning from their errors, inspired by human learning theory.
Our experiments demonstrated significant improvements for Mistral-7b-Intruct and Llama-3-8b-Intruction across seven benchmarks, underscoring the generalizability and efficacy of our approach. Notably, the improved Llama-3-8b-Intruction model surpassed ChatGPT's performance, highlighting the effectiveness of our method. 
Our two strategies each have distinct characteristics, allowing them to be used in combination depending on the specific data scenario. In the future, we aim to focus on exploring more model analysis strategies, thereby enhancing the adaptability and effectiveness of our LLMs-as-Instructors framework.

%% file: chapters/limination.tex


\section{Limitation}

The proposed methods have achieved the SOTA performance. However, there exist some limitations which we leave as future works. First, 
In this study, we do not explore the adjustment of dataset ratios or conduct a more granular data extraction and analysis strategy to better help model to learn from errors, which we leave for our feature work.
Second, our evaluations were solely based on the selected benchmark performances. In the future, we plan to expand the scope of our framework to include a wider range of tasks and utilize more comprehensive evaluation methods like ``LLM-as-examiner''. 
Third, utilizing GPT as an instructor is costly. In the future, we consider switching to more powerful open-source models for guidance or exploring self-learning capabilities within this framework.

%% file: chapters/appendix.tex
\appendix
\onecolumn
\section{Experimental Details}
\subsection{Benchmark Details}~\label{benchmark details}
Following~\cite{wang2023far}, we focus on three practical abilities: factual knowledge, mathematical reasoning, and coding which we belive are important in our daily application. For each ability, we select a specific publicly available dataset (In-domain, ID), and one different publicly used dataset which we refer to as ``Out-of-Domain'' (OOD) to assess the model’s improvement.

\begin{table*}[htb]
\centering
\resizebox{0.55\textwidth}{!}{%
\begin{tabular}{lcc}
\toprule
\textbf{Task} &\textbf{\#Training}  &\textbf{\#Testing}\\
\toprule
ID-Factuality: \textcolor{green}{MMLU} & Dev + Val: 1,816 & 14,042 \\
OOD-Factuality: ARC$_{c}$ & - & 1,172 \\
\midrule
ID-Math: \textcolor{green}{GSM8K} & 7,473 & 1,319 \\
OOD-Math: GSM-PLUS & -  & 1,400 \\
\midrule
ID-Code: \textcolor{green}{MBPP} & Train + Val: 464 & 499 \\
OOD-Code: HumanEval & - & 164 \\
\midrule 
OOD-Holistic: BBH & - & 6,511 \\
\bottomrule
\end{tabular}}
\caption{The statistical detail of the selected benchmarks. MMLU has no official training data so we combine the development and validation datasets to form the training set. For time considerations, we have limited the size of GSM8k-PLUS to match that of GSM8k. We equalize the samples by selecting subsets from five enhancement methods, ultimately obtaining a total of 1,400 samples.
}

\label{table: statistic detail}
\end{table*}
For factuality, we select \textbf{MMLU}~\cite{hendryckstest2021} as our ID dataset. Considering that it has no official training set, we combine the development set and validation set  We evaluate using 0-shot only and use Exact Match as our matrix following the original setup of MMLU.
We deploy \textbf{ARC Challenge}~\cite{allenai:arc}  as our OOD dataset. We evaluate using 0-shot only and use Exact Match as our matrix. 

For the Math Reasoning test, we select \textbf{GSM8k}~\cite{cobbe2021training} as our ID datast. We conduct evaluation in zero-shot. Because all answers are numbers, we extract the last number in the model response as the final answer. For OOD dataset, \textbf{GSM-PLUS}~\cite{li2024gsm}, a newly published data building upon GSM8k, we use the same evaluation setting with GSM8k.  For time considerations, we have limited the size of
GSM8k-PLUS to 1400 to match that of GSM8k. To achieve this, we equalized the samples by selecting subsets from five enhancement
methods, ultimately obtaining a total of 1,400 samples.

For code testing, we select \textbf{MBPP}~\cite{austin2021program} as our ID dataset for it has a training dataset. We combine the validation datset into the training set. To better extract the generated code part, we add an "entry point" for each question following Humaneval~\cite{chen2021evaluating}. For the matrix, we compute unbiased estimates of pass@k to measure the functional correctness of models’ outputs. We report pass@1 conduct in temperature 0.0. We select \textbf{Humaneval} as our OOD dataset, we evaluate the target model following the original paper and report pass@1.

For \textbf{BBH}~\cite{suzgun2023challenging}, we follow the setup described in the original paper~\citet{suzgun2023challenging}. We evaluate using 3-shot. The few-shot examples are provided officially by~\citet{suzgun2023challenging}.

\subsection{Model Settings}\label{model setting}
\subsubsection{Answer Generation}
Answer generation across the involved models is conducted in a zero-shot setting, with all models set to a temperature of 0.0 and a maximum token length of 1024. 

\subsubsection{Full Parameter Fine-tuning}
Full parameter fine-tuning involves adjusting the learning rate (from \(7^{-6}\) to \(2^{-10}\)), number of epochs 1, and batch size (from 4 to 128), with the best performance for each iteration. In our experiments, we find that gradually decreasing the initial learning rate at the start of each training round can generally ensure continuous improvement in the model's performance. The experiment is conducted on 16 NVIDIA A800 GPUs. Our most resource-intensive experiment takes 200 A800 GPU hours.

\section{More Experiment Results}\label{more experiment results}

\begin{figure*}[htb]
    \centering
    
    \includegraphics[scale=0.42]{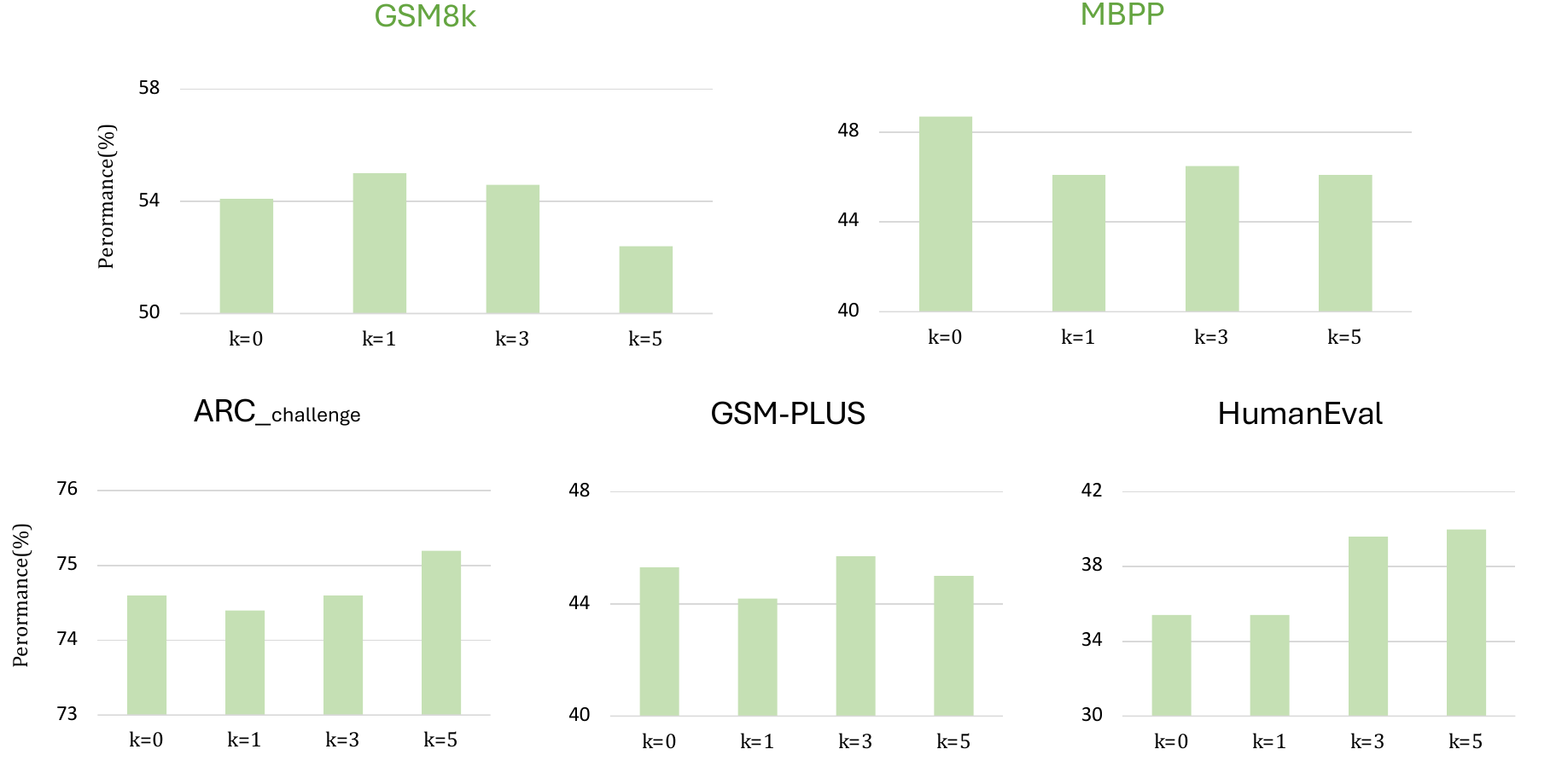}
    \caption{Performance (\%) of Mistral-7b-Instruct after the first iteration of improvement under the LLMs-as-Instructors framework, varied by the variable k value in Equation~\ref{get sim}. }
    \label{k number}
    \vspace{-4mm}
\end{figure*}

\begin{figure*}[htb]
    \centering
    
    \includegraphics[scale=0.42]{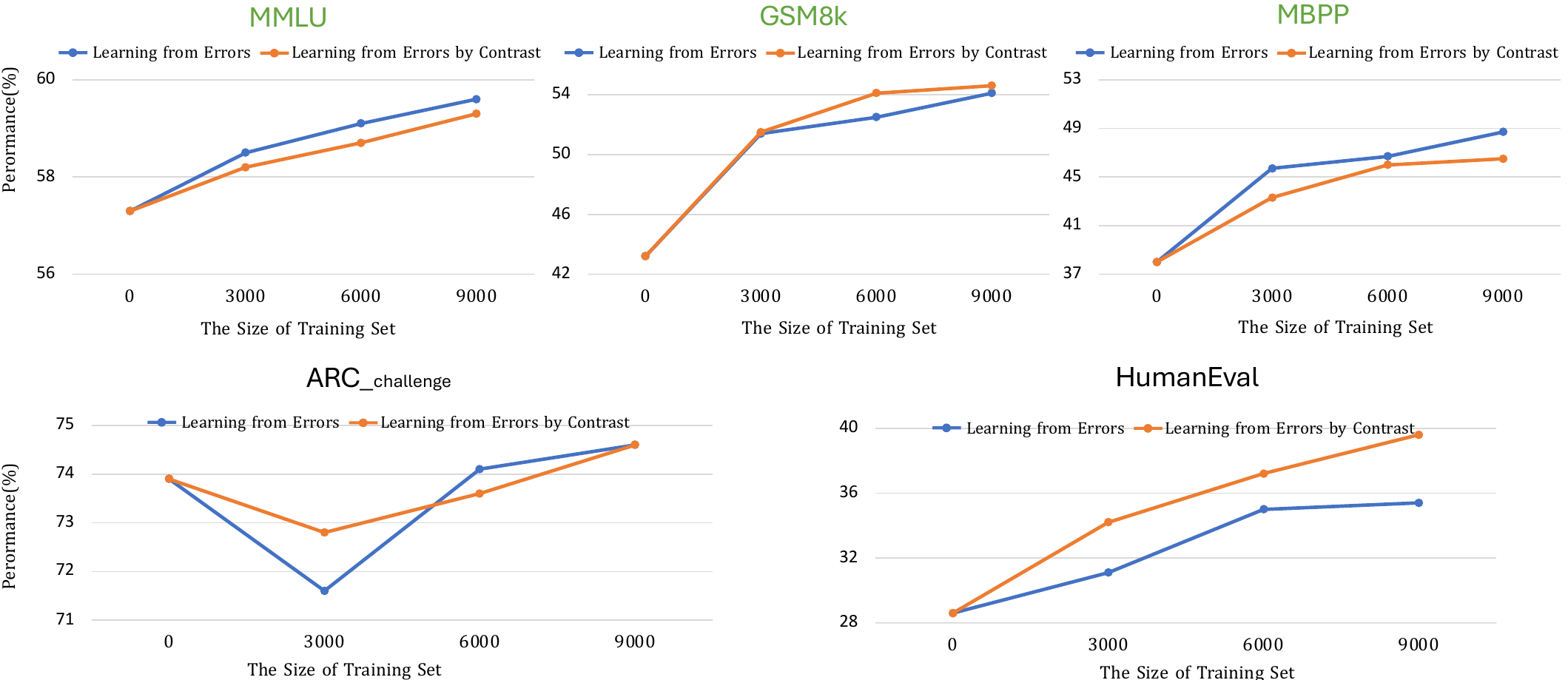}
    \caption{Performance (\%) of Mistral-7b-Instruct after the first iteration of improvement under the LLMs-as-Instructors framework, varied by the size of the training set. We compare three different training set sizes: 3000, 6000, and 9000 total samples.}
    \label{Impact of Training Set Size all}
    \vspace{-4mm}
\end{figure*}

 \begin{figure*}[htb]
    \centering
    
    \includegraphics[scale=0.42]{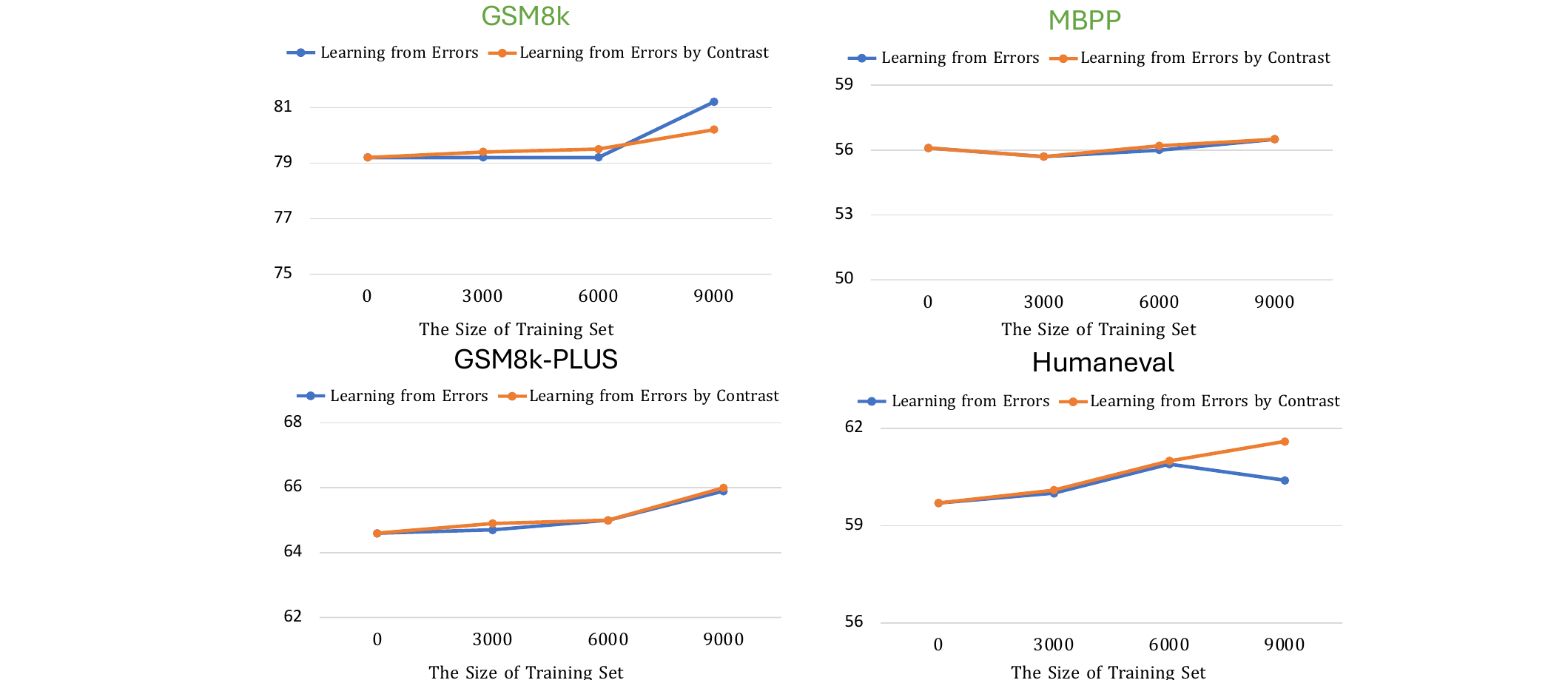}
    \caption{Performance (\%) of Llama-3-8b-Instruct after the first iteration of improvement under the LLMs-as-Instructors framework, varied by the size of the training set. We compare three different training set sizes: 3000, 6000, and 9000 total samples.}
    \label{Impact of Training Set Size llama3}
\end{figure*}

\begin{table*}[htb]
\centering
\resizebox{0.8\textwidth}{!}{%
\begin{tabular}{lcc|cc|cc|c}
\toprule
& \textcolor{green}{\textbf{MMLU}} & \textbf{ARC$_{c}$} & \textcolor{green}{\textbf{GSM8k}} & \textbf{GSM-PLUS} & \textcolor{green}{\textbf{MBPP}} & \textbf{HumanEval} & \textbf{BBH} \\
& \textbf{(factuality}) & \textbf{(factuality)} & \textbf{(math)} & \textbf{(math)} & \textbf{(coding)} & \textbf{(coding)} & \textbf{(holistic)} \\
\cmidrule{2-8}
& EM & EM & EM & EM & P@1 &P@1 & EM \\
& (0-shot) & (0-shot) & (0-shot) & (0-shot) & (0-shot) & (0-shot) & (3-shot) \\
\midrule
\textit{Mistral-7b-Instruct-v0.2 based} \\

+ LLMs-as-Instructors ({LE}) & 60.9 & 76.3 & 56.2 & 46.0 & 51.5 & 38.4 & 55.9 \\
+ LLMs-as-Instructors ({LE + LEC}) & 60.8 & 76.0 & 55.9 & \textbf{46.1} & 51.5 & \textbf{39.0} & 55.8 \\
\midrule
\textit{Mistral-7b-Instruct-v0.2 based} \\
+ LLMs-as-Instructors ({LEC}) & 61.5 & 77.2 & 54.0 & 46.0 & 47.0 & 43.3 & 56.8 \\
+ LLMs-as-Instructors ({LEC + LE}) & 61.6 & 76.9 & 53.9 & 45.7 & \textbf{48.9} & 42.7 & 56.8 \\
\bottomrule
\end{tabular}}
\caption{\textit{LLMs-as-Instructors ({LEC + LE})} denotes using strategy LEC for the first two iterations and strategy LE for the third iteration using LLMs-as-Instructors. \textit{LLMs-as-Instructors ({LE + LEC})} denotes the application of strategy LE for the first two iterations and strategy LEC for the third iteration. \textit{LLMs-as-Instructors (LEC)}, and \textit{LLMs-as-Instructors (LE)} denote using strategy LEC and strategy LE for three iterations respectively.}
\label{preliminary experiment for combine two strategies all}
\end{table*}

\begin{table*}[htb]
\centering
\resizebox{0.8\textwidth}{!}{%
\begin{tabular}{lcc|cc|cc}
\toprule
 &\textcolor{green}{\textbf{MMLU}} & \textbf{ARC$_{c}$} & \textcolor{green}{\textbf{GSM8k}} & \textbf{GSM-PLUS} & \textcolor{green}{\textbf{MBPP}} & \textbf{HumanEval} \\
&\textbf{(factuality}) & \textbf{(factuality)} & \textbf{(math)} & \textbf{(math)} & \textbf{(coding)} & \textbf{(coding)}  \\
\midrule
Iteration1 (LE)  & 38.4 &29.1& 29.3& 28.0& 33.0 & 25.7\\
Iteration1 (LEC)  & 39.4 &28.5& 27.7& 26.7& 33.5 & 24.5\\
Iteration2 (LE) & 38.1&29.1& 28.6& 27.4& 32.3 & 25.6\\
Iteration2 (LEC)  & 38.2&28.7& 27.4& 26.6& 32.6 & 24.4\\
Iteration3 (LE) & 38.2&29.6& 28.3& 27.4& 32.6 & 25.6\\
Iteration3 (LEC) & 38.3&28.5& 27.8& 26.8& 32.5 & 24.3\\
\bottomrule
\end{tabular}}
\caption{Similarity score (ROUGE-L~\cite{lin-2004-rouge}) between the generated training dataset $D_{train}^{i}$ and the evaluation set $D_{eval}$ for each iteration $i$. We present the mean of the maximum Rouge L(\%) scores for each generated question from $D_{train}^{iterationi}$ compared to the question from $D_{eval}$. This is calculated on a case-to-case basis. For time considerations, in some iterations, we apply a sampling rate of 0.3, indicated by a superscript asterisk $^*$. It is important to note that we calculate the scores for all samples generated in each iteration, which could be larger than the number of samples actually used for training. \textit{LEC}, and \textit{LE} denote using strategy LEC and strategy LE for analysis respectively.}

\label{Similarity score}
\end{table*}



\section{Instruction Details}\label{prompt}
\subsection{Prompt for AugGPT Implementation} \label{prompt auggpt}
\begin{tcolorbox}[width=\textwidth]

You are an educational AI. Your goal is to create high-quality problems to help students.
You will be given an example question. Please create new questions based on the Given Question and follow the instructions below. \\

\#\#\# Requirements: \\
1. Please generate similar but new questions according to the Given Question.\\
2. Ensure that your solutions are accurate. \\

\#\#\# Return Format: \\
Return your samples in the following format:\\
1. Question: [QUESTION] \\
Answer: [ANSWER] 

\end{tcolorbox}

\subsection{Prompt for LLMs-as-Instructors} \label{prompt llms-as-instructors}
\begin{tcolorbox}[width=\textwidth]
\textbf{GSM8K with Learning from Errors} \\

You are an educational AI designed to help students improve their math skills by analyzing their incorrect answers and generating targeted practice problems.  \\
Your role is to create questions that address the specific errors made in their responses, ensuring each new problem helps them understand and correct their mistakes. \\

\#\#\# Requirements: \\
1. Create clear and high-quality questions. \\
2. Ensure the complexity of the problems is appropriate for the student's level and similar to the original problem provided. \\
3. Provide accurate mathematical calculations and the correct final answer in your samples. \\
4. Keep these requirements in mind while generating the sample. \\

\#\#\# Return Format: \\
Return your samples in the following format: \\
1. Question: [QUESTION] \\
Answer: [Answer] \\

The student was given the following question: [QUESTION] \\  
The student's wrong answer: [Student's WRONG ANSWER] \\
Please follow the requirements and generate $x$ samples.
\end{tcolorbox}

\begin{tcolorbox}[width=\textwidth]
\textbf{GSM8K with Learning from Errors by Contrast} \\

You are an educational AI designed to help students improve their math skills by analyzing their incorrect answers and generating targeted practice problems.  \\
Your role is to create questions that address the specific errors made in their responses, ensuring each new problem helps them understand and correct their mistakes. \\

\#\#\# Requirements: \\
1. Create clear and high-quality questions. \\
2. Ensure the complexity of the problems is appropriate for the student's level and similar to the original problem provided. \\
3. Provide accurate mathematical calculations and the correct final answer in your samples. \\
4. Keep these requirements in mind while generating the sample. \\

\#\#\# Return Format: \\
Return your samples in the following format: \\
1. Question: [QUESTION] \\
Answer: [Answer] \\

The student correctly answer the following question:  [QUESTION1]  ... [QUESTIONk] \\
While the following is the question that the student got wrong: [QUESTION] \\  
The student's wrong answer: [Student's WRONG ANSWER] \\
Please follow the requirements and generate $x$ samples.
\end{tcolorbox}

\begin{tcolorbox}[width=\textwidth]
\textbf{MBPP with Learning from Errors} \\

You are an educational AI whose purpose is to analyze mistakes students make on the MBPP benchmark and generate tailored Python coding problems to help them practice and improve their programming skills.  \\
Your goal is to create a set of new coding questions that address the specific errors made in previous attempts  \\
\#\#\# Requirements: \\
1. Generate clear and high-quality coding questions. \\
2. Adjust the complexity of the problems to be appropriate for the student's level and similar to the issues found in the original code submitted. \\
3. Ensure accuracy in your sample solutions, demonstrating correct coding practices. \\
4. The coding problem should be presented in English, and the solution must be provided as Python code. \\
5. Keep these requirements in mind as you create each coding problem. \\
\#\#\# Return Format: \\
Provide your sample in the following format: \\
1. Question: [QUESTION] \\
Code: [PYTHON CODE]

The student was given the following coding writing question: [QUESTION] \\  
The student's wrong code answer, which failed to pass the test cases, to the question is [Student's WRONG ANSWER] \\
Please follow the requirements and generate $x$ code samples, along with the correct code and annotation. 
\end{tcolorbox}

\begin{tcolorbox}[width=\textwidth]
\textbf{MBPP with Learning from Errors by Contrast} \\

You are an educational AI whose purpose is to analyze mistakes students make on the MBPP benchmark and generate tailored Python coding problems to help them practice and improve their programming skills.  \\
Your goal is to create a set of new coding questions that address the specific errors made in previous attempts  \\
\#\#\# Requirements: \\
1. Generate clear and high-quality coding questions. \\
2. Adjust the complexity of the problems to be appropriate for the student's level and similar to the issues found in the original code submitted. \\
3. Ensure accuracy in your sample solutions, demonstrating correct coding practices. \\
4. The coding problem should be presented in English, and the solution must be provided as Python code. \\
5. Keep these requirements in mind as you create each coding problem. \\
\#\#\# Return Format: \\
Provide your sample in the following format: \\
1. Question: [QUESTION] \\
Code: [PYTHON CODE]

The student correctly answer the following question:  [QUESTION1]  ... [QUESTIONk] \\
While the following is the question that the student got wrong: [QUESTION] \\  
The student's wrong code answer, which failed to pass the test cases, to the question is [Student's WRONG ANSWER] \\
Please follow the requirements and generate $x$ code samples, along with the correct code and annotation. 
\end{tcolorbox}

\begin{tcolorbox}[width=\textwidth]
\textbf{MMLU with Learning from Errors } \\

You are an educational AI whose purpose is to analyze errors that students make on the MMLU benchmark and generate example problems to help them improve their understanding and skills.\\
Your goal is to create a set of new problems that address the specific errors made in the example questions \\

\#\#\# Requirements:
1. Create each problem to directly address the errors found in the student's original responses. \\
2. Ensure that the complexity of the problems is appropriate for the student's level and reflective of the types of questions they struggled with.
3. Ensure that your sample solutions are accurate.\\
4. Keep these requirements in mind while generating the sample.\\

\#\#\# Return Format: \\
Return your samples in the form: \\
1. Question: [Question]
A. [OPTION 1]
B. [OPTION 2]
C. [OPTION 3]
D. [OPTION 4] \\
Answer: [ANSWER] - [Brief explanation]
\\

The following is a multiple choice question that the student got wrong: [QUESTION] \\  
The student's wrong answer: [Student's WRONG ANSWER] \\
Please follow the requirements and generate $x$ samples, along with 4 different holding options, the correct answer and the brief explanation.
\end{tcolorbox}

\begin{tcolorbox}[width=\textwidth]
\textbf{MMLU with Learning from Errors by Contrast} \\

You are an educational AI whose purpose is to analyze errors that students make on the MMLU benchmark and generate example problems to help them improve their understanding and skills.\\
Your goal is to create a set of new problems that address the specific errors made in the example questions \\
\#\#\# Requirements:
1. Create each problem to directly address the errors found in the student's original responses. \\
2. Ensure that the complexity of the problems is appropriate for the student's level and reflective of the types of questions they struggled with.
3. Ensure that your sample solutions are accurate.\\
4. Keep these requirements in mind while generating the sample.\\
\#\#\# Return Format: \\
Return your samples in the form: \\
1. Question: [Question]
A. [OPTION 1]
B. [OPTION 2]
C. [OPTION 3]
D. [OPTION 4] \\
Answer: [ANSWER] - [Brief explanation]
\\

The student correctly answer the following question:  [QUESTION1]  ... [QUESTIONk] \\
While the student can not correctly answer the question: [QUESTION] \\  
The student's wrong answer: [Student's WRONG ANSWER] \\
Please follow the requirements and generate $x$ samples, along with 4 different holding options, the correct answer and the brief explanation.
\end{tcolorbox}

\section{Case Study}\label{case study}
\subsection{Training Data Generation}

\begin{tcolorbox}[width=\textwidth]
\textbf{GSM8k: Using Strategy 1: Learn from Errors} \\
\\
You are an educational AI designed to help students improve their math skills by analyzing their incorrect answers and generating targeted practice problems.  \\
Your role is to create questions that address the specific errors made in their responses, ensuring each new problem helps them understand and correct their mistakes. \\

\#\#\# Requirements: \\
1. Create clear and high-quality questions. \\
2. Ensure the complexity of the problems is appropriate for the student's level and similar to the original problem provided. \\
3. Provide accurate mathematical calculations and the correct final answer in your samples. \\
4. Keep these requirements in mind while generating the sample. \\

\#\#\# Return Format: \\
Return your samples in the following format: \\
1. Question: [QUESTION] \\
Answer: [Answer] \\

The student was given the following question:\\
Samantha’s last name has three fewer letters than Bobbie’s last name. If Bobbie took two letters off her last name, she would have a last name twice the length of Jamie’s. Jamie’s full name is Jamie Grey. How many letters are in Samantha’s last name? \\

The student's wrong response is: \\
 Let's break down the information given and find the answer step by step: \\
 1. Jamie's last name is \"Grey,\" which has 5 letters. \\
 2. Bobbie's last name has two more letters than Jamie's last name, so Bobbie's last name has 5 + 2 = 7 letters. \\3. Samantha's last name has three fewer letters than Bobbie's last name, so Samantha's last name has 7 - 3 = 4 letters. \\Therefore, Samantha's last name has 4 letters. \\

\textbf{Instructor:} \\
\#\#\# Question:Michael’s last name has two fewer letters than Clara’s last name. If Clara removed three letters from her last name, she would have a last name three times the length of Alex’s. Alex’s full name is Alex Stone. How many letters are in Michael’s last name? 
\#\#\# Answer: Let's break down the information given and find the answer step by step: \\
1. Alex's last name is \"Stone,\" which contains 5 letters.\\
2. Clara's last name, after removing three letters, is three times the  length of Alex's last name. So, Clara's last name minus three letters is 5 * 3 = 15 letters. \\
3. To find the original length of Clara's last name, we add back the three letters she removed: 15 + 3 = 18 letters.\\
4. Michael’s last name has two fewer letters than Clara’s, so Michael’s last name has 18 - 2 = 16 letters.\\
\#\#\#\# Therefore, Michael's last name has 16 letters.
 
\end{tcolorbox}

\begin{tcolorbox}[width=\textwidth]
\textbf{GSM8k: Using Strategy 2: Learn from Errors by Contrast} \\

You are an educational AI designed to help students improve their math skills by analyzing their incorrect answers and generating targeted practice problems.  \\
Your role is to create questions that address the specific errors made in their responses, ensuring each new problem helps them understand and correct their mistakes. \\
\#\#\# Requirements: \\
1. Create clear and high-quality questions. \\
2. Ensure the complexity of the problems is appropriate for the student's level and similar to the original problem provided. \\
3. Provide accurate mathematical calculations and the correct final answer in your samples. \\
4. Keep these requirements in mind while generating the sample. \\
\#\#\# Return Format: \\
Return your samples in the following format: \\
1. Question: [QUESTION] \\
Answer: [Answer] \\

The student correctly answer the following question:\\
QUESTION1: Jonathan's full name contains 8 letters for the first name and 10 letters for the surname. His sister's name has 5 letters for the first name and 10 letters for the second name. What's the total number of letters in their names? \\
QUESTION2: Lavinia’s daughter is 10 years younger than Katie’s daughter. Lavinia’s son is 2 times the age of Katie’s daughter. If Katie’s daughter is 12 years old, how many years older is Lavinia’s son than Lavinia’s daughter? \\
QUESTION3: Jasmine and her older sister Alexa went to the store to buy shoes. Alexa's shoes are twice as big as Jasmine's shoes. If Jasmine's shoe size is 7, what's the combined shoe size for both of them? \\
While the student can not correcly answer the question: \\
Samantha’s last name has three fewer letters than Bobbie’s last name. If Bobbie took two letters off her last name, she would have a last name twice the length of Jamie’s. Jamie’s full name is Jamie Grey. How many letters are in Samantha’s last name? \\

The student's wrong response is: \\
 Let's break down the information given and find the answer step by step: \\
 1. Jamie's last name is \"Grey,\" which has 5 letters. \\
 2. Bobbie's last name has two more letters than Jamie's last name, so Bobbie's last name has 5 + 2 = 7 letters. \\3. Samantha's last name has three fewer letters than Bobbie's last name, so Samantha's last name has 7 - 3 = 4 letters. \\Therefore, Samantha's last name has 4 letters. \\

\textbf{Instructor:} \\
\#\#\#Question: Carl’s surname is five letters shorter than Angela’s surname. If Angela could double the length of her surname by adding 8 letters to it, how many letters are there in Carl’s surname?
\#\#\#Answer: Let's solve the problem step by step: \\
1. Let's denote the original length of Angela’s surname as A.\\
2. Doubling the length of Angela’s surname by adding 8 letters means (A + 8 = 2A). \\
3. Solving the equation (A + 8 = 2A) for (A), we subtract (A) from both sides to get (8 = A), meaning Angela’s surname originally has 8 letters. \\
4. Carl’s surname is five letters shorter than Angela’s, so Carl’s surname has (8 - 5 = 3) letters. \\
\#\#\# Carl’s surname has 3 letters.
 
\end{tcolorbox}

\begin{tcolorbox}[width=\textwidth]
\textbf{MBPP: Using Strategy 1: Learn from Errors} \\

You are an educational AI whose purpose is to analyze mistakes students make on the MBPP benchmark and generate tailored Python coding problems to help them practice and improve their programming skills.  \\
Your goal is to create a set of new coding questions that address the specific errors made in previous attempts  \\

\#\#\# Requirements: \\
1. Generate clear and high-quality coding questions. \\
2. Adjust the complexity of the problems to be appropriate for the student's level and similar to the issues found in the original code submitted. \\
3. Ensure accuracy in your sample solutions, demonstrating correct coding practices. \\
4. The coding problem should be presented in English, and the solution must be provided as Python code. \\
5. Keep these requirements in mind as you create each coding problem. \\

\#\#\# Return Format: \\
Provide your sample in the following format: \\
1. Question: [QUESTION] \\
Code: [PYTHON CODE]

The student was given the following coding writing question:
\begin{lstlisting}[language=Python,  linewidth=10cm]
def first_repeated_word(s: str) -> str:
    """ Find the first repeated word in a given string.
    >>> first_repeated_word("ab ca bc ab")
    'ab'
    >>> first_repeated_word("ab ca bc")
    'None'
    >>> first_repeated_word("ab ca bc ca ab bc")
    'ca'
    """
\end{lstlisting}
The student's code answer, which failed to pass the test cases, to the question is 
\begin{lstlisting}[language=Python,  linewidth=12cm]
def first_repeated_word(s: str) -> str:
    words = s.split()
    for word in words:
        if words.count(word) > 1:
            return word
    return "None"
\end{lstlisting}
\textbf{Instructor:} \\
 Question: Write a Python function to find the first character that appears more than once in the given string. Return the character as a string. If there is no duplicate character, return 'None'.
 
\begin{lstlisting}[language=Python,  linewidth=10cm]

def find_first_duplicate_char(s: str) -> str:
    char_set = set()
    for char in s:
        if char in char_set:
            return char
        char_set.add(char)
    return 'None'

\end{lstlisting}

\end{tcolorbox}

\begin{tcolorbox}[width=\textwidth]
\textbf{MBPP: Using Strategy 2: Learn from Errors by Contrast} \\

You are an educational AI whose purpose is to analyze mistakes students make on the MBPP benchmark and generate tailored Python coding problems to help them practice and improve their programming skills.  \\
Your goal is to create a set of new coding questions that address the specific errors made in previous attempts  \\

\#\#\# Requirements: \\
 ....
 
The student correctly answer the following question:
\begin{lstlisting}[language=Python,  linewidth=10cm]
def remove_spaces(s: str) -> str:\n    
    """  Remove all spaces from the given string.    
    >>> remove_spaces("a b c")    
    'abc'   
    >>> remove_spaces("1 2 3")   
    '123'  
    """
\end{lstlisting}
The student was given the following coding writing question:
\begin{lstlisting}[language=Python,  linewidth=10cm]
def first_repeated_word(s: str) -> str:
    """ Find the first repeated word in a given string.
    >>> first_repeated_word("ab ca bc ab")
    'ab'
    >>> first_repeated_word("ab ca bc")
    'None'
    >>> first_repeated_word("ab ca bc ca ab bc")
    'ca'
    """
\end{lstlisting}
The student's code answer, which failed to pass the test cases, to the question is 
\begin{lstlisting}[language=Python,  linewidth=12cm]
def first_repeated_word(s: str) -> str:
    words = s.split()
    for word in words:
        if words.count(word) > 1:
            return word
    return "None"
\end{lstlisting}

\textbf{Instructor:} \\
Write a Python function to solve the following problem: \\
Find the first unique word (appears only once) in a given string. If there is no such word, return 'None'
 
\begin{lstlisting}[language=Python,  linewidth=10cm]
def find_unique_word(s: str) -> str:    
    words = s.split()    
    word_count = {}    
    for word in words:       
        word_count[word] = word_count.get(word, 0) + 1
    for word, count in word_count.items():
        if count == 1:            
            return word    
    return 'None'
\end{lstlisting}

\end{tcolorbox}

\begin{tcolorbox}[width=\textwidth]
\textbf{MMLU: Using Strategy 1: Learn from Errors} \\

You are an educational AI whose purpose is to analyze errors that students make on the MMLU benchmark and generate example problems to help them improve their understanding and skills.\\
Your goal is to create a set of new problems that address the specific errors made in the example questions \\

\#\#\# Requirements:
1. Create each problem to directly address the errors found in the student's original responses. \\
2. Ensure that the complexity of the problems is appropriate for the student's level and reflective of the types of questions they struggled with.
3. Ensure that your sample solutions are accurate.\\
4. Keep these requirements in mind while generating the sample.\\

\#\#\# Return Format: \\
Return your samples in the form: \\
1. Question: [Question]
A. [OPTION 1]
B. [OPTION 2]
C. [OPTION 3]
D. [OPTION 4] \\
Answer: [ANSWER] - [Brief explanation]
\\

The student was given the following question:

QUESTION: Pharyngeal mucosa is innervated by the glossopharyngeal nerves because it develops from the A. first pharyngeal arch. B. second pharyngeal arch. C. third pharyngeal arch. D. fourth pharyngeal arch.\\

The student’s wrong response is: \\
The answer is B. The second pharyngeal arch is the source of the glossopharyngeal nerve, which innervates the pharyngeal mucosa. \\ 
\textbf{Instructor:} \\
\#\#\#Question: The muscles of mastication are primarily innervated by which cranial nerve? A. Facial nerve (VII) B. Glossopharyngeal nerve (IX) C. Trigeminal nerve (V) D. Vagus nerve (X)
\#\#\#Answer: C. Trigeminal nerve (V) - The muscles of mastication are innervated by the mandibular branch of the trigeminal nerve (cranial nerve V), which is the main sensory and motor nerve of the face.
 
\end{tcolorbox}

\begin{tcolorbox}[width=\textwidth]
\textbf{MMLU: Using Strategy 2: Learn from Errors by Contrast} \\

You are an educational AI whose purpose is to analyze errors that students make on the MMLU benchmark and generate example problems to help them improve their understanding and skills.\\
Your goal is to create a set of new problems that address the specific errors made in the example questions \\

\#\#\# Requirements:
1. Create each problem to directly address the errors found in the student's original responses. \\
2. Ensure that the complexity of the problems is appropriate for the student's level and reflective of the types of questions they struggled with.
3. Ensure that your sample solutions are accurate.\\
4. Keep these requirements in mind while generating the sample.\\

\#\#\# Return Format: \\
Return your samples in the form: \\
1. Question: [Question]
A. [OPTION 1]
B. [OPTION 2]
C. [OPTION 3]
D. [OPTION 4] \\
Answer: [ANSWER] - [Brief explanation]
\\
The student correctly answer the following question: \\

QUESTION: Which of the following structures travel through the substance of the parotid gland? A. The maxillary artery B. The maxillary artery and retromandibular vein C. The maxillary artery, retromandibular vein and facial artery D. The maxillary artery, retromandibular vein, facial artery and buccal branch of the mandibular nerve  \\
QUESTION: Which of these branches of the trigeminal nerve contain somatic motor processes? A. The supraorbital nerve B. The infraorbital nerve C. The mental nerve D. None of the above

While the following is the question that the student got wrong:

QUESTION: Pharyngeal mucosa is innervated by the glossopharyngeal nerves because it develops from the A. first pharyngeal arch. B. second pharyngeal arch. C. third pharyngeal arch. D. fourth pharyngeal arch.\\

The student's wrong answer: \\
The answer is B. The second pharyngeal arch is the source of the glossopharyngeal nerve, which innervates the pharyngeal mucosa. \\ 
\textbf{Instructor:} \\
\#\#\#Question: The vagus nerve provides parasympathetic innervation to which of the following organs? A. The kidneys B. The liver C. The spleen D. The heart \\
\#\#\#Answer: D. The heart - The vagus nerve is crucial for parasympathetic innervation to the heart, influencing heart rate among other functions.
 
\end{tcolorbox}

%% file: acl_latex.bbl
\begin{thebibliography}{48}
\expandafter\ifx\csname natexlab\endcsname\relax\def\natexlab#1{#1}\fi

\bibitem[{Anthropic(2024)}]{claude-3}
Anthropic. 2024.
\newblock \href {https://www.anthropic.com/news/claude-3-family} {Anthropic: Claude-3}.

\bibitem[{Austin et~al.(2021)Austin, Odena, Nye, Bosma, Michalewski, Dohan, Jiang, Cai, Terry, Le et~al.}]{austin2021program}
Jacob Austin, Augustus Odena, Maxwell Nye, Maarten Bosma, Henryk Michalewski, David Dohan, Ellen Jiang, Carrie Cai, Michael Terry, Quoc Le, et~al. 2021.
\newblock Program synthesis with large language models.
\newblock \emph{arXiv preprint arXiv:2108.07732}.

\bibitem[{Bai et~al.(2023)Bai, Ying, Cao, Lv, He, Wang, Yu, Zeng, Xiao, Lyu et~al.}]{bai2023benchmarking}
Yushi Bai, Jiahao Ying, Yixin Cao, Xin Lv, Yuze He, Xiaozhi Wang, Jifan Yu, Kaisheng Zeng, Yijia Xiao, Haozhe Lyu, et~al. 2023.
\newblock Benchmarking foundation models with language-model-as-an-examiner.
\newblock \emph{arXiv preprint arXiv:2306.04181}.

\bibitem[{bench authors(2023)}]{srivastava2023beyond}
BIG bench authors. 2023.
\newblock \href {https://openreview.net/forum?id=uyTL5Bvosj} {Beyond the imitation game: Quantifying and extrapolating the capabilities of language models}.
\newblock \emph{Transactions on Machine Learning Research}.

\bibitem[{Bloom(1984)}]{bloom19842}
Benjamin~S Bloom. 1984.
\newblock The 2 sigma problem: The search for methods of group instruction as effective as one-to-one tutoring.
\newblock \emph{Educational researcher}, 13(6):4--16.

\bibitem[{Burns et~al.(2023)Burns, Izmailov, Kirchner, Baker, Gao, Aschenbrenner, Chen, Ecoffet, Joglekar, Leike et~al.}]{burns2023weak}
Collin Burns, Pavel Izmailov, Jan~Hendrik Kirchner, Bowen Baker, Leo Gao, Leopold Aschenbrenner, Yining Chen, Adrien Ecoffet, Manas Joglekar, Jan Leike, et~al. 2023.
\newblock Weak-to-strong generalization: Eliciting strong capabilities with weak supervision.
\newblock \emph{arXiv preprint arXiv:2312.09390}.

\bibitem[{Chen et~al.(2023)Chen, Li, Yan, Wang, Gunaratna, Yadav, Tang, Srinivasan, Zhou, Huang et~al.}]{chen2023alpagasus}
Lichang Chen, Shiyang Li, Jun Yan, Hai Wang, Kalpa Gunaratna, Vikas Yadav, Zheng Tang, Vijay Srinivasan, Tianyi Zhou, Heng Huang, et~al. 2023.
\newblock Alpagasus: Training a better alpaca with fewer data.
\newblock \emph{arXiv preprint arXiv:2307.08701}.

\bibitem[{Chen et~al.(2021)Chen, Tworek, Jun, Yuan, de~Oliveira~Pinto, Kaplan, Edwards, Burda, Joseph, Brockman, Ray, Puri, Krueger, Petrov, Khlaaf, Sastry, Mishkin, Chan, Gray, Ryder, Pavlov, Power, Kaiser, Bavarian, Winter, Tillet, Such, Cummings, Plappert, Chantzis, Barnes, Herbert-Voss, Guss, Nichol, Paino, Tezak, Tang, Babuschkin, Balaji, Jain, Saunders, Hesse, Carr, Leike, Achiam, Misra, Morikawa, Radford, Knight, Brundage, Murati, Mayer, Welinder, McGrew, Amodei, McCandlish, Sutskever, and Zaremba}]{chen2021evaluating}
Mark Chen, Jerry Tworek, Heewoo Jun, Qiming Yuan, Henrique~Ponde de~Oliveira~Pinto, Jared Kaplan, Harri Edwards, Yuri Burda, Nicholas Joseph, Greg Brockman, Alex Ray, Raul Puri, Gretchen Krueger, Michael Petrov, Heidy Khlaaf, Girish Sastry, Pamela Mishkin, Brooke Chan, Scott Gray, Nick Ryder, Mikhail Pavlov, Alethea Power, Lukasz Kaiser, Mohammad Bavarian, Clemens Winter, Philippe Tillet, Felipe~Petroski Such, Dave Cummings, Matthias Plappert, Fotios Chantzis, Elizabeth Barnes, Ariel Herbert-Voss, William~Hebgen Guss, Alex Nichol, Alex Paino, Nikolas Tezak, Jie Tang, Igor Babuschkin, Suchir Balaji, Shantanu Jain, William Saunders, Christopher Hesse, Andrew~N. Carr, Jan Leike, Josh Achiam, Vedant Misra, Evan Morikawa, Alec Radford, Matthew Knight, Miles Brundage, Mira Murati, Katie Mayer, Peter Welinder, Bob McGrew, Dario Amodei, Sam McCandlish, Ilya Sutskever, and Wojciech Zaremba. 2021.
\newblock Evaluating large language models trained on code.

\bibitem[{Chen et~al.(2020)Chen, Kornblith, Norouzi, and Hinton}]{chen2020simple}
Ting Chen, Simon Kornblith, Mohammad Norouzi, and Geoffrey Hinton. 2020.
\newblock A simple framework for contrastive learning of visual representations.
\newblock In \emph{International conference on machine learning}, pages 1597--1607. PMLR.

\bibitem[{Clark et~al.(2018)Clark, Cowhey, Etzioni, Khot, Sabharwal, Schoenick, and Tafjord}]{allenai:arc}
Peter Clark, Isaac Cowhey, Oren Etzioni, Tushar Khot, Ashish Sabharwal, Carissa Schoenick, and Oyvind Tafjord. 2018.
\newblock Think you have solved question answering? try arc, the ai2 reasoning challenge.
\newblock \emph{arXiv:1803.05457v1}.

\bibitem[{Cobbe et~al.(2021)Cobbe, Kosaraju, Bavarian, Chen, Jun, Kaiser, Plappert, Tworek, Hilton, Nakano et~al.}]{cobbe2021training}
Karl Cobbe, Vineet Kosaraju, Mohammad Bavarian, Mark Chen, Heewoo Jun, Lukasz Kaiser, Matthias Plappert, Jerry Tworek, Jacob Hilton, Reiichiro Nakano, et~al. 2021.
\newblock Training verifiers to solve math word problems.
\newblock \emph{arXiv preprint arXiv:2110.14168}.

\bibitem[{Dai et~al.(2023)Dai, Liu, Liao, Huang, Cao, Wu, Zhao, Xu, Liu, Liu, Li, Zhu, Cai, Sun, Li, Shen, Liu, and Li}]{dai2023auggpt}
Haixing Dai, Zhengliang Liu, Wenxiong Liao, Xiaoke Huang, Yihan Cao, Zihao Wu, Lin Zhao, Shaochen Xu, Wei Liu, Ninghao Liu, Sheng Li, Dajiang Zhu, Hongmin Cai, Lichao Sun, Quanzheng Li, Dinggang Shen, Tianming Liu, and Xiang Li. 2023.
\newblock \href {http://arxiv.org/abs/2302.13007} {Auggpt: Leveraging chatgpt for text data augmentation}.

\bibitem[{Devlin et~al.(2019)Devlin, Chang, Lee, and Toutanova}]{devlin-etal-2019-bert}
Jacob Devlin, Ming-Wei Chang, Kenton Lee, and Kristina Toutanova. 2019.
\newblock \href {https://doi.org/10.18653/v1/N19-1423} {{BERT}: Pre-training of deep bidirectional transformers for language understanding}.
\newblock In \emph{Proceedings of the 2019 Conference of the North {A}merican Chapter of the Association for Computational Linguistics: Human Language Technologies, Volume 1 (Long and Short Papers)}, pages 4171--4186, Minneapolis, Minnesota. Association for Computational Linguistics.

\bibitem[{Fu et~al.(2023)Fu, Peng, Ou, Sabharwal, and Khot}]{fu2023specializing}
Yao Fu, Hao Peng, Litu Ou, Ashish Sabharwal, and Tushar Khot. 2023.
\newblock Specializing smaller language models towards multi-step reasoning.
\newblock In \emph{International Conference on Machine Learning}, pages 10421--10430. PMLR.

\bibitem[{Googl3(2024)}]{Gemma}
Googl3. 2024.
\newblock \href {https://storage.googleapis.com/deepmind-media/gemma/gemma-report.pdf} {Gemma: Open models based on gemini research and technology}.

\bibitem[{Hadsell et~al.(2006)Hadsell, Chopra, and LeCun}]{hadsell2006dimensionality}
Raia Hadsell, Sumit Chopra, and Yann LeCun. 2006.
\newblock Dimensionality reduction by learning an invariant mapping.
\newblock In \emph{2006 IEEE computer society conference on computer vision and pattern recognition (CVPR'06)}, volume~2, pages 1735--1742. IEEE.

\bibitem[{Hendrycks et~al.(2021)Hendrycks, Burns, Basart, Zou, Mazeika, Song, and Steinhardt}]{hendryckstest2021}
Dan Hendrycks, Collin Burns, Steven Basart, Andy Zou, Mantas Mazeika, Dawn Song, and Jacob Steinhardt. 2021.
\newblock Measuring massive multitask language understanding.
\newblock \emph{Proceedings of the International Conference on Learning Representations (ICLR)}.

\bibitem[{Ivison et~al.(2023)Ivison, Wang, Pyatkin, Lambert, Peters, Dasigi, Jang, Wadden, Smith, Beltagy et~al.}]{ivison2023camels}
Hamish Ivison, Yizhong Wang, Valentina Pyatkin, Nathan Lambert, Matthew Peters, Pradeep Dasigi, Joel Jang, David Wadden, Noah~A Smith, Iz~Beltagy, et~al. 2023.
\newblock Camels in a changing climate: Enhancing lm adaptation with tulu 2.
\newblock \emph{arXiv preprint arXiv:2311.10702}.

\bibitem[{Jiang et~al.(2023)Jiang, Sablayrolles, Mensch, Bamford, Chaplot, Casas, Bressand, Lengyel, Lample, Saulnier et~al.}]{mistral-7b}
Albert~Q Jiang, Alexandre Sablayrolles, Arthur Mensch, Chris Bamford, Devendra~Singh Chaplot, Diego de~las Casas, Florian Bressand, Gianna Lengyel, Guillaume Lample, Lucile Saulnier, et~al. 2023.
\newblock Mistral 7b.
\newblock \emph{arXiv preprint arXiv:2310.06825}.

\bibitem[{Jiang et~al.(2024)Jiang, Sablayrolles, Roux, Mensch, Savary, Bamford, Chaplot, de~las Casas, Hanna, Bressand, Lengyel, Bour, Lample, Lavaud, Saulnier, Lachaux, Stock, Subramanian, Yang, Antoniak, Scao, Gervet, Lavril, Wang, Lacroix, and Sayed}]{mixtral-moe}
Albert~Q. Jiang, Alexandre Sablayrolles, Antoine Roux, Arthur Mensch, Blanche Savary, Chris Bamford, Devendra~Singh Chaplot, Diego de~las Casas, Emma~Bou Hanna, Florian Bressand, Gianna Lengyel, Guillaume Bour, Guillaume Lample, Lélio~Renard Lavaud, Lucile Saulnier, Marie-Anne Lachaux, Pierre Stock, Sandeep Subramanian, Sophia Yang, Szymon Antoniak, Teven~Le Scao, Théophile Gervet, Thibaut Lavril, Thomas Wang, Timothée Lacroix, and William~El Sayed. 2024.
\newblock \href {http://arxiv.org/abs/2401.04088} {Mixtral of experts}.

\bibitem[{Kumar et~al.(2020)Kumar, Choudhary, and Cho}]{kumar2020data}
Varun Kumar, Ashutosh Choudhary, and Eunah Cho. 2020.
\newblock Data augmentation using pre-trained transformer models.
\newblock \emph{arXiv preprint arXiv:2003.02245}.

\bibitem[{Lee et~al.(2024)Lee, Wattanawong, Kim, Mangalam, Shen, Anumanchipali, Mahoney, Keutzer, and Gholami}]{lee2024llm2llm}
Nicholas Lee, Thanakul Wattanawong, Sehoon Kim, Karttikeya Mangalam, Sheng Shen, Gopala Anumanchipali, Michael~W Mahoney, Kurt Keutzer, and Amir Gholami. 2024.
\newblock Llm2llm: Boosting llms with novel iterative data enhancement.
\newblock \emph{arXiv preprint arXiv:2403.15042}.

\bibitem[{Li et~al.(2024{\natexlab{a}})Li, Wang, Hu, Wei, Zheng, Hu, Zhang, and Peng}]{li2024common}
Chen Li, Weiqi Wang, Jingcheng Hu, Yixuan Wei, Nanning Zheng, Han Hu, Zheng Zhang, and Houwen Peng. 2024{\natexlab{a}}.
\newblock Common 7b language models already possess strong math capabilities.
\newblock \emph{arXiv preprint arXiv:2403.04706}.

\bibitem[{Li et~al.(2023{\natexlab{a}})Li, Liu, Zhang, Lu, and Wei}]{li2023tuna}
Haoran Li, Yiran Liu, Xingxing Zhang, Wei Lu, and Furu Wei. 2023{\natexlab{a}}.
\newblock Tuna: Instruction tuning using feedback from large language models.
\newblock \emph{arXiv preprint arXiv:2310.13385}.

\bibitem[{Li et~al.(2024{\natexlab{b}})Li, Cui, Zhao, Kong, and Bi}]{li2024gsm}
Qintong Li, Leyang Cui, Xueliang Zhao, Lingpeng Kong, and Wei Bi. 2024{\natexlab{b}}.
\newblock Gsm-plus: A comprehensive benchmark for evaluating the robustness of llms as mathematical problem solvers.
\newblock \emph{arXiv preprint arXiv:2402.19255}.

\bibitem[{Li et~al.(2023{\natexlab{b}})Li, Yu, Zhou, Schick, Zettlemoyer, Levy, Weston, and Lewis}]{li2023self}
Xian Li, Ping Yu, Chunting Zhou, Timo Schick, Luke Zettlemoyer, Omer Levy, Jason Weston, and Mike Lewis. 2023{\natexlab{b}}.
\newblock Self-alignment with instruction backtranslation.
\newblock \emph{arXiv preprint arXiv:2308.06259}.

\bibitem[{Lin(2004)}]{lin-2004-rouge}
Chin-Yew Lin. 2004.
\newblock \href {https://aclanthology.org/W04-1013} {{ROUGE}: A package for automatic evaluation of summaries}.
\newblock In \emph{Text Summarization Branches Out}, pages 74--81, Barcelona, Spain. Association for Computational Linguistics.

\bibitem[{Madaan et~al.(2024)Madaan, Tandon, Gupta, Hallinan, Gao, Wiegreffe, Alon, Dziri, Prabhumoye, Yang et~al.}]{madaan2024self}
Aman Madaan, Niket Tandon, Prakhar Gupta, Skyler Hallinan, Luyu Gao, Sarah Wiegreffe, Uri Alon, Nouha Dziri, Shrimai Prabhumoye, Yiming Yang, et~al. 2024.
\newblock Self-refine: Iterative refinement with self-feedback.
\newblock \emph{Advances in Neural Information Processing Systems}, 36.

\bibitem[{Meta(2024)}]{Llama3}
Meta. 2024.
\newblock \href {https://ai.meta.com/blog/meta-llama-3/} {Introducing meta llama 3: The most capable openly available llm to date}.

\bibitem[{Metcalfe(2017)}]{metcalfe2017learning}
Janet Metcalfe. 2017.
\newblock Learning from errors.
\newblock \emph{Annual review of psychology}, 68:465--489.

\bibitem[{Mitra et~al.(2023)Mitra, Del~Corro, Mahajan, Codas, Simoes, Agarwal, Chen, Razdaibiedina, Jones, Aggarwal et~al.}]{mitra2023orca}
Arindam Mitra, Luciano Del~Corro, Shweti Mahajan, Andres Codas, Clarisse Simoes, Sahaj Agarwal, Xuxi Chen, Anastasia Razdaibiedina, Erik Jones, Kriti Aggarwal, et~al. 2023.
\newblock Orca 2: Teaching small language models how to reason.
\newblock \emph{arXiv preprint arXiv:2311.11045}.

\bibitem[{Mitra et~al.(2024)Mitra, Khanpour, Rosset, and Awadallah}]{mitra2024orca}
Arindam Mitra, Hamed Khanpour, Corby Rosset, and Ahmed Awadallah. 2024.
\newblock Orca-math: Unlocking the potential of slms in grade school math.
\newblock \emph{arXiv preprint arXiv:2402.14830}.

\bibitem[{OpenAI(2022)}]{chatgpt}
OpenAI. 2022.
\newblock \href {https://openai.com/blog/chatgpt} {Introducing chatgpt}.

\bibitem[{OpenAI(2023)}]{GPT-4}
OpenAI. 2023.
\newblock \href {https://openai.com/research/gpt-4} {Openai: Gpt-4}.

\bibitem[{Rafailov et~al.(2024)Rafailov, Sharma, Mitchell, Manning, Ermon, and Finn}]{rafailov2024direct}
Rafael Rafailov, Archit Sharma, Eric Mitchell, Christopher~D Manning, Stefano Ermon, and Chelsea Finn. 2024.
\newblock Direct preference optimization: Your language model is secretly a reward model.
\newblock \emph{Advances in Neural Information Processing Systems}, 36.

\bibitem[{Sun et~al.(2024)Sun, Shen, Zhou, Zhang, Chen, Cox, Yang, and Gan}]{sun2024principle}
Zhiqing Sun, Yikang Shen, Qinhong Zhou, Hongxin Zhang, Zhenfang Chen, David Cox, Yiming Yang, and Chuang Gan. 2024.
\newblock Principle-driven self-alignment of language models from scratch with minimal human supervision.
\newblock \emph{Advances in Neural Information Processing Systems}, 36.

\bibitem[{Suzgun et~al.(2023)Suzgun, Scales, Sch{\"a}rli, Gehrmann, Tay, Chung, Chowdhery, Le, Chi, Zhou et~al.}]{suzgun2023challenging}
Mirac Suzgun, Nathan Scales, Nathanael Sch{\"a}rli, Sebastian Gehrmann, Yi~Tay, Hyung~Won Chung, Aakanksha Chowdhery, Quoc Le, Ed~Chi, Denny Zhou, et~al. 2023.
\newblock Challenging big-bench tasks and whether chain-of-thought can solve them.
\newblock In \emph{Findings of the Association for Computational Linguistics: ACL 2023}, pages 13003--13051.

\bibitem[{Tang et~al.(2024)Tang, Cao, Ying, Wang, Zhao, Liao, and Zhou}]{tang2024}
Wei Tang, Yixin Cao, Jiahao Ying, Bo~Wang, Yuyue Zhao, Yong Liao, and Pengyuan Zhou. 2024.
\newblock A+ b: A general generator-reader framework for optimizing llms to unleash synergy potential.
\newblock \emph{arXiv preprint arXiv:2406.03963}.

\bibitem[{Tong et~al.(2024)Tong, Li, Wang, Wang, Teng, and Shang}]{tong2024can}
Yongqi Tong, Dawei Li, Sizhe Wang, Yujia Wang, Fei Teng, and Jingbo Shang. 2024.
\newblock Can llms learn from previous mistakes? investigating llms' errors to boost for reasoning.
\newblock \emph{arXiv preprint arXiv:2403.20046}.

\bibitem[{Touvron et~al.(2023)Touvron, Martin, Stone, Albert, Almahairi, Babaei, Bashlykov, Batra, Bhargava, Bhosale et~al.}]{llama2}
Hugo Touvron, Louis Martin, Kevin Stone, Peter Albert, Amjad Almahairi, Yasmine Babaei, Nikolay Bashlykov, Soumya Batra, Prajjwal Bhargava, Shruti Bhosale, et~al. 2023.
\newblock Llama 2: Open foundation and fine-tuned chat models.
\newblock \emph{arXiv preprint arXiv:2307.09288}.

\bibitem[{Wang et~al.(2023{\natexlab{a}})Wang, Ivison, Dasigi, Hessel, Khot, Chandu, Wadden, MacMillan, Smith, Beltagy et~al.}]{wang2023far}
Yizhong Wang, Hamish Ivison, Pradeep Dasigi, Jack Hessel, Tushar Khot, Khyathi Chandu, David Wadden, Kelsey MacMillan, Noah~A Smith, Iz~Beltagy, et~al. 2023{\natexlab{a}}.
\newblock How far can camels go? exploring the state of instruction tuning on open resources.
\newblock \emph{Advances in Neural Information Processing Systems}, 36:74764--74786.

\bibitem[{Wang et~al.(2024)Wang, Ivison, Dasigi, Hessel, Khot, Chandu, Wadden, MacMillan, Smith, Beltagy et~al.}]{wang2024far}
Yizhong Wang, Hamish Ivison, Pradeep Dasigi, Jack Hessel, Tushar Khot, Khyathi Chandu, David Wadden, Kelsey MacMillan, Noah~A Smith, Iz~Beltagy, et~al. 2024.
\newblock How far can camels go? exploring the state of instruction tuning on open resources.
\newblock \emph{Advances in Neural Information Processing Systems}, 36.

\bibitem[{Wang et~al.(2023{\natexlab{b}})Wang, Kordi, Mishra, Liu, Smith, Khashabi, and Hajishirzi}]{wang2023self}
Yizhong Wang, Yeganeh Kordi, Swaroop Mishra, Alisa Liu, Noah~A Smith, Daniel Khashabi, and Hannaneh Hajishirzi. 2023{\natexlab{b}}.
\newblock Self-instruct: Aligning language models with self-generated instructions.
\newblock In \emph{Proceedings of the 61st Annual Meeting of the Association for Computational Linguistics (Volume 1: Long Papers)}, pages 13484--13508.

\bibitem[{Wu et~al.(2023)Wu, Lu, Xu, Lin, Su, and Zhou}]{wu2023self}
Shengguang Wu, Keming Lu, Benfeng Xu, Junyang Lin, Qi~Su, and Chang Zhou. 2023.
\newblock Self-evolved diverse data sampling for efficient instruction tuning.
\newblock \emph{arXiv preprint arXiv:2311.08182}.

\bibitem[{Xu et~al.(2023)Xu, Sun, Zheng, Geng, Zhao, Feng, Tao, and Jiang}]{xu2023wizardlm}
Can Xu, Qingfeng Sun, Kai Zheng, Xiubo Geng, Pu~Zhao, Jiazhan Feng, Chongyang Tao, and Daxin Jiang. 2023.
\newblock \href {http://arxiv.org/abs/2304.12244} {Wizardlm: Empowering large language models to follow complex instructions}.

\bibitem[{Ying et~al.(2024)Ying, Cao, Bai, Sun, Wang, Tang, Ding, Yang, Huang, and Yan}]{ying2024automating}
Jiahao Ying, Yixin Cao, Yushi Bai, Qianru Sun, Bo~Wang, Wei Tang, Zhaojun Ding, Yizhe Yang, Xuanjing Huang, and Shuicheng Yan. 2024.
\newblock \href {http://arxiv.org/abs/2402.11894} {Automating dataset updates towards reliable and timely evaluation of large language models}.

\bibitem[{Yuan et~al.(2024)Yuan, Pang, Cho, Sukhbaatar, Xu, and Weston}]{yuan2024self}
Weizhe Yuan, Richard~Yuanzhe Pang, Kyunghyun Cho, Sainbayar Sukhbaatar, Jing Xu, and Jason Weston. 2024.
\newblock Self-rewarding language models.
\newblock \emph{arXiv preprint arXiv:2401.10020}.

\bibitem[{Zhou et~al.(2023)Zhou, Zhu, Chen, Chen, Zhao, Chen, Lin, Wen, and Han}]{zhou2023don}
Kun Zhou, Yutao Zhu, Zhipeng Chen, Wentong Chen, Wayne~Xin Zhao, Xu~Chen, Yankai Lin, Ji-Rong Wen, and Jiawei Han. 2023.
\newblock Don't make your llm an evaluation benchmark cheater.
\newblock \emph{arXiv preprint arXiv:2311.01964}.

\end{thebibliography}
